\documentclass[conf]{new-aiaa}
\usepackage[utf8]{inputenc}

\usepackage{graphicx}
\usepackage{amsmath}
\usepackage{bm}
\usepackage[version=4]{mhchem}
\usepackage{siunitx}
\usepackage{longtable,tabularx}
\usepackage{subcaption}
\usepackage{adjustbox}
\usepackage{afterpage}
\usepackage{pdflscape}

\DeclareMathOperator*{\argmax}{arg\,max}
\DeclareMathOperator*{\argmin}{arg\,min}
\newcommand{\hvec}[1]{\underline{\bm{\mathrm{#1}}}}
\newcommand{\vvec}[1]{\bm{\mathrm{#1}}}
\newcommand{\oB}{\mathrm{B}}
\newcommand{\oC}{\mathrm{C}}
\newcommand{\oS}{\mathrm{S}}
\newcommand{\oI}{\mathrm{I}}
\newcommand{\fB}{\mathcal{B}}
\newcommand{\fC}{\mathcal{C}}
\newcommand{\fS}{\mathcal{S}}

\newcommand{\desk}[1]{ \left[#1\right]^\vee}
\newcommand{\given}[1][]{\:#1\vert\:}
\newcommand{\sk}[1]{  \left[#1\right]^\wedge }
\newcommand{\Exp}[1]{\exp{\left( \sk{#1}\right)}}

\usepackage{colortbl} 
\usepackage{booktabs} 
\newcommand{\ra}[1]{\renewcommand{\arraystretch}{#1}} 
\usepackage{xcolor}
\usepackage{tikz}
\usetikzlibrary{shapes,arrows,calc,patterns,angles,quotes,fadings,decorations.pathreplacing}
\usepackage{tikz-3dplot}
\usepackage{tkz-graph}
\tikzstyle{vertex}=[circle, draw, inner sep=0pt, minimum size=8pt]
\newcommand{\vertex}{\node[vertex]}

\tikzset{%
  >=latex, 
  inner sep=0pt,%
  outer sep=1pt,%
  mark coordinate/.style={inner sep=0pt,outer sep=0pt,minimum size=3pt,
    fill=black,circle}%
}

\definecolor{desert}{rgb}{0.76, 0.6, 0.42}
\definecolor{darkpastelgreen}{rgb}{0.01, 0.75, 0.24}
\definecolor{darkspringgreen}{rgb}{0.09, 0.45, 0.27}
\definecolor{cobalt}{rgb}{0.0, 0.28, 0.67}
\definecolor{darkgreen}{rgb}{0.0, 0.5, 0.0}
\definecolor{coquelicot}{rgb}{1.0, 0.22, 0.0}
\definecolor{canaryyellow}{rgb}{1.0, 0.94, 0.0}
\definecolor{cadetgrey}{rgb}{0.57, 0.64, 0.69}
\definecolor{cadet}{rgb}{0.33, 0.41, 0.47}
\definecolor{mediumcandyapplered}{rgb}{0.89, 0.02, 0.17}
\definecolor{dimgray}{rgb}{0.41, 0.41, 0.41}

\title{Keypoint-based Stereophotoclinometry for Characterizing and Navigating Small Bodies: A Factor Graph Approach}

\author{Travis Driver\footnote{PhD Candidate, Institute for Robotics and Intelligent Machines, School of Aerospace Engineering, Georgia Institute of Technology, Atlanta, GA.}, Andrew Vaughan\footnote{Senior Engineer, Mission Design and Navigation, Jet Propulsion Laboratory, California Institute of Technology, Pasadena, CA}, Yang Cheng\footnote{Robotics Technologist, Mobility and Robotic Systems, Jet Propulsion Laboratory, California Institute of Technology, Pasadena, CA.}, Adnan Ansar\footnote{Group Supervisor,  Mobility and Robotic Systems, Jet Propulsion Laboratory, California Institute of Technology, Pasadena, CA.}, John Christian\footnote{Associate Professor, School of Aerospace Engineering, Georgia Institute of Technology, Atlanta, GA.}, and Panagiotis Tsiotras\footnote{David \& Andrew Lewis Chair, Professor, Institute for Robotics and Intelligent Machines, School of Aerospace Engineering, Georgia Institute of Technology, Atlanta, GA.}}

\begin{document}

\maketitle


\begin{abstract}
This paper proposes the incorporation of techniques from stereophotoclinometry (SPC) into a keypoint-based structure-from-motion (SfM) system to estimate the surface normal and albedo at detected landmarks to improve \textit{autonomous} surface and shape characterization of small celestial bodies from \textit{in-situ} imagery. 
In contrast to the current state-of-the-practice method for small body shape reconstruction, i.e., SPC, which relies on human-in-the-loop verification and high-fidelity \textit{a priori} information to achieve accurate results, we forego the expensive maplet estimation step and instead leverage dense keypoint measurements and correspondences from an \textit{autonomous} keypoint detection and matching method based on deep learning to provide the necessary photogrammetric constraints. 
Moreover, we develop a factor graph-based approach allowing for \textit{simultaneous} optimization of the spacecraft's pose, landmark positions, Sun-relative direction, and surface normals and albedos via fusion of Sun sensor measurements and image keypoint measurements. 
The proposed framework is validated on \textit{real} imagery of the Cornelia crater on Asteroid 4 Vesta, along with pose estimation and mapping comparison against an SPC reconstruction, where we demonstrate precise alignment to the SPC solution without relying on any \textit{a priori} camera pose and topography information or humans-in-the-loop.
\end{abstract}






\section{Introduction}
\vspace{5pt}
\lettrine{T}here has been an increasing interest in missions to small bodies (e.g., asteroids, comets) due to their great scientific value, with four currently in operation (OSIRIS-REx, Hayabusa2, Lucy, Psyche) and three scheduled to launch over the next five years (Hera, EMA, Janus). 
In addition to planetary protection~\cite{cheng2018} and resource utilization~\cite{mazanek2015,rivkin2019}, small bodies are believed to be remnants of the solar system's formation, and studying their composition could provide insight into the evolution of the solar system and the origins of organic life on Earth~\cite{barucci2011}. 
These missions currently rely on an extended characterization phase, where a shape model is reconstructed from images acquired during a ground-controlled trajectory around the body, as shape models are essential for characterizing the body and estimating the spacecraft's relative pose in subsequent phases~\cite{bhaskaran2011}. 
However, current state-of-the-practice shape reconstruction methods rely on humans-in-the-loop and accurate \textit{a priori} information to ensure accurate results. 

\begin{figure}[t!]
    \centering
    \includegraphics[width=0.8\linewidth]{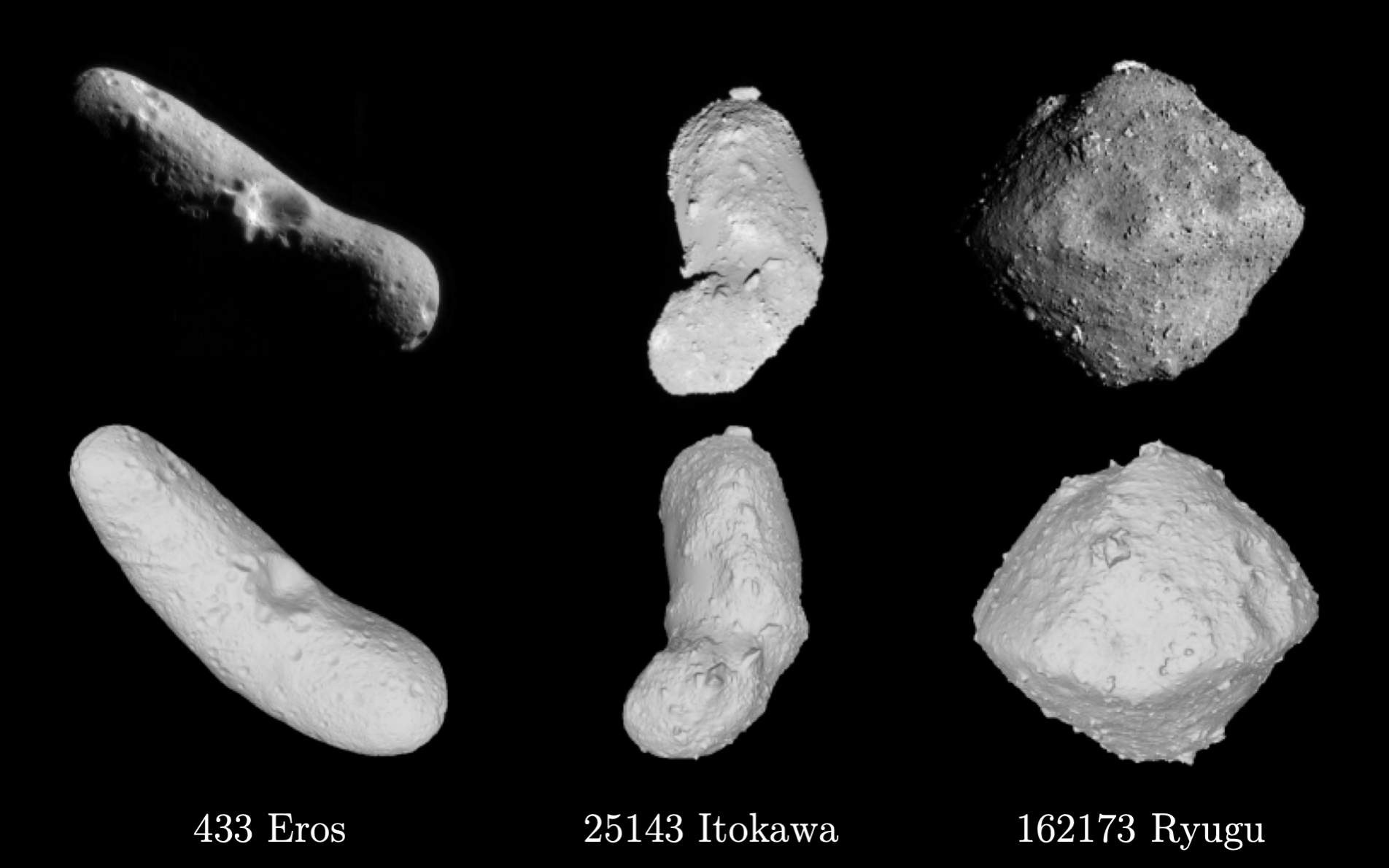}
    \caption{Examples of shape models produced by SPC.}
    \label{fig:spc-examples}
\end{figure}

Stereophotoclinometry~\cite{gaskell2008} (SPC) is the current method of choice for 3D reconstruction of small bodies and has been used to model a broad suite of celestial bodies, including the Moon, 433 Eros, 25143 Itokawa~\cite{gaskell2008}, 4 Vesta~\cite{raymond2011, mastrodemos2012}, and 101955 Bennu~\cite{lorenz2017, barnouin2020} (see Figure \ref{fig:spc-examples}).
While SPC has proven effective, the process requires extensive human-in-the-loop verification and high-fidelity \textit{a priori} information to achieve accurate results. 
Specifically, SPC attempts to estimate a collection of digital terrain maps (DTMs), high-resolution local topography and albedo maps, through direct alignment of ortho-rectified projections, or \textit{orthoimages}, of a given surface patch from multiple images predicated on an initial shape model and accurate \textit{a priori} estimates of the spacecraft's pose (position and orientation). 
Photometric stereo techniques are applied to derive surface gradients and albedos of the imaged surface patches at each pixel of the orthoimages. 
The local topography solution is fixed upon convergence, typically requiring human input to achieve precise alignment to the images, and used to refine pose and landmark position estimates through a multistep iterative process by rendering the DTM and aligning it across multiple views~\cite{gaskell2023psj}. 
Finally, the local DTMs can be collated into a global shape model by exploiting overlap and limb constraints within a separate iterative processing loop. 
While this approach has achieved much success, its reliance on extensive human involvement for extended durations and a complex multistep optimization process limits mission capabilities and increases operational costs~\cite{quadrelli2015,nesnas2021,getzandanner2022scitech}. 

In contrast to the traditional maplet-based approach of SPC, this work proposes a keypoint-based framework that leverages dense image keypoint measurements and correspondences within a structure-from-motion (SfM) system. 
Indeed, SfM and simultaneous localization and mapping (SLAM), which leverage autonomous keypoint detection and matching methods to estimate correspondences between images, have been shown to be promising technologies for \textit{autonomous} optical navigation for missions to small bodies~\cite{dor2021cvpr,dor2022astroslam}. 
Consequently, we propose the incorporation of photometric stereo constraints and Sun vector measurements into a graph-based SfM system to estimate surface normals and albedos at estimated landmarks, providing detailed information for surface characterization and shape reconstruction. 
The proposed framework, which leverages \textit{factor graphs}~\cite{dellaert2017} to model and solve the complex SPC estimation problem, forgoes the expensive local maplet estimation step, streamlines the optimization process, and renders SPC more amenable to recent advances in computer vision, including feature detection, description, and matching methods based on \textit{deep learning}~\cite{driver2022astrovision} and graph-based incremental smoothing~\cite{kaess2012}. 
Specifically, the contributions of this work are as follows:

\begin{enumerate}
    \item we propose the fusion of measurements from onboard monocular cameras and Sun sensors in a SfM system to reconstruct a landmark map of the body and the relative pose of the spacecraft while \textit{simultaneously} estimating the surface normal and albedo at landmarks; 
    \item we model the photometric stereo process of SPC using the formalism of \textit{factor graphs} and analyze its performance with respect to the chosen reflectance model;
    \item we leverage keypoint measurements and correspondences from a state-of-the-art, \textit{autonomous} keypoint detection and matching method based on deep learning~\cite{edstedt2023dkm} and apply the proposed framework to \textit{real} imagery of the Cornelia crater on Asteroid 4 Vesta, demonstrating precise alignment to an SPC derived map.
\end{enumerate}


\section{Preliminaries}
\vspace{5pt}
In this section, we include some relevant mathematical concepts and notation.
Specifically, we summarize the representation of 3D poses as elements of the Special Euclidean group $\mathrm{SE}(3)$ (Section \ref{sec:rigid-transformations}), and introduce a minimal representation of unit three-vectors (Section \ref{sec:unit-vectors}).


\subsection{3D Rigid Body Transformations} \label{sec:rigid-transformations}

We represent the relative position and orientation of the spacecraft, that is, its relative \textit{pose}, as an element of the Special Euclidean group $\mathrm{SE}(3)$ by a matrix
\begin{equation}
    T_{\fB\fS} \triangleq
    \begin{bmatrix}
    R_{\fB\fS} & \vvec{r}^\fB_{\oS\oB} \\
    \bm{0}_{1 \times 3}      & 1
    \end{bmatrix}, 
\end{equation}
where $R_{\fB\fS} \in \mathrm{SO}(3)$ is the orientation of some body-fixed frame of the small body $\fB$ with respect to a spacecraft body-fixed frame $\fS$, and $\vvec{r}^\fB_{\oS\oB} \in \mathbb{R}^3$ is the position of the spacecraft's origin with respect to the origin of $\fB$, expressed in $\fB$. 
Moreover, the fixed pose $T_{\fS\fC}$ of the onboard camera relative to the body-fixed frame of the spacecraft is precisely known \textit{a priori}, which can be used to derive the sequence of \textit{the camera's relative pose} $T_{\fB \fC_k} = T_{\fB \fS_k}T_{\fS_k \fC_k} = T_{\fB \fS_k}T_{\fS\fC}$ at each time index $k$. 


The uncertainty of the camera's pose, using the exponential map (at the identity) of the $\mathrm{SO}(3)$ group of rotations and the \textit{hat} operator (see~\cite{chirikjian2011}), can be defined as
\begin{align}
	T_{\fB\fC} \triangleq 
	\overline{T}_{\fB\fC} 
	\begin{bmatrix}
		\Exp{\vvec{\omega}} & \vvec{\nu} \\ 
		\vvec{0}_{1 \times 3}             & 1
	\end{bmatrix},
\end{align}
where $\vvec{\omega} \sim \mathcal{N}\left( \vvec{0}_{3 \times 1}, \Sigma_R\right)$, $\vvec{\nu} \sim \mathcal{N}\left( \vvec{0}_{3 \times 1}, \Sigma_r\right)$, and $\overline{T}_{\fB\fS}$ is the \textit{actual} pose~\cite{dellaert2017,forster2016manifold}. 
Therefore, the estimated pose $T_{\fB\fC_k}$ is represented by the uncertain orientation $R_{\fB\fC_k} \triangleq \overline{R}_{\fB\fC_k}\Exp{\bm{\omega}}$ and the uncertain position $\vvec{r}^{\fB}_{\oC_k\oB} \triangleq \overline{\vvec{r}}_{\oC_k\oB}^{\fB} + \overline{R}_{\fB\fC_k}\vvec{\nu}$. 
This relates to a \textit{retraction} $\mathcal{R}_T(\vvec{\gamma}, \vvec{\tau})$ (at a pose $T\in\mathrm{SE}(3)$) that maps the local coordinates $\vvec{\zeta} = [\vvec{\gamma}^\top\, \vvec{\tau}^\top]^\top \in \mathbb{R}^6$ to an element of $\mathrm{SE}(3)$ via
\begin{align} \label{eq:ret-pose}
	\mathcal{R}_T(\vvec{\gamma}, \vvec{\tau}) \triangleq 
	T
	\begin{bmatrix}
		\Exp{\vvec{\gamma}} & \vvec{\tau} \\ 
		\vvec{0}_{1 \times 3}             & 1
	\end{bmatrix} =
        \begin{bmatrix}
		R\Exp{\vvec{\gamma}} & \vvec{r} + R\vvec{\tau} \\ 
		\vvec{0}_{1 \times 3}             & 1
	\end{bmatrix}.
\end{align}
Thus, the optimization of elements of $\mathrm{SE}(3)$ may be framed in terms of the coordinates $\vvec{\zeta} \in \mathbb{R}^6$~\cite{forster2016manifold,absil2007trust}.


\subsection{The Unit 2-Sphere} \label{sec:unit-vectors}

An important two-dimensional manifold is the unit 2-sphere $\mathbb{S}^2 \triangleq \left\{ \vvec{x} \in \mathbb{R}^3 \given \|\vvec{x}\| = 1\right\}$, i.e., the topological space composed of all unit vectors in $\mathbb{R}^3$.
The tangent space $\mathcal{T}_{\vvec{x}}\left(\mathbb{S}^2\right)$ at a point $\vvec{x} \in \mathbb{S}^2$ is defined as the set of all three-vectors tangent to $\mathbb{S}^2$ at $\vvec{x}$, i.e., 
\begin{equation}
    \mathcal{T}_{\vvec{x}}\left(\mathbb{S}^2\right) \triangleq \left\{\vvec{y} \in \mathbb{R}^3 \given \vvec{x}^\top\vvec{y} = 0, \vvec{x} \in \mathbb{S}^2\right\}.
\end{equation}
For any $\vvec{y} \in \mathcal{T}_{\vvec{x}}\left(\mathbb{S}^2\right)$, we can write $\vvec{y} = B_{\vvec{x}}\vvec{\xi}$ with $\vvec{\xi} \in \mathbb{R}^2$ in the plane tangent to $\mathbb{S}^2$ at $\vvec{x}$ defined by the basis $B_{\vvec{x}} \in \mathbb{R}^{3 \times 2}$. 
With these definitions, we can define another useful retraction $\mathcal{R}_{\vvec{x}}(\vvec{\xi})$ as follows~\cite{dellaert2017}:
\begin{equation} \label{eq:ret-unit}
    \mathcal{R}_{\vvec{x}}(\vvec{\xi}) \triangleq \cos(\|B_{\vvec{x}}\vvec{\xi}\|)\vvec{x} + \sin(\|B_{\vvec{x}}\vvec{\xi}\|)\frac{B_{\vvec{x}}\vvec{\xi}}{\|B_{\vvec{x}}\vvec{\xi}\|} \in \mathbb{S}^2.
\end{equation}
This minimal representation allows for the optimization of the unit vector $\vvec{x} \in \mathbb{S}^2$ with respect to the local coordinates $\vvec{\xi} \in \mathbb{R}^2$ according to the basis $B_{\vvec{x}}$. 
Uncertainty in the unit vector can also be defined in the local coordinate system defined by $B_{\overline{\vvec{x}}}$ at the true value $\overline{\vvec{x}}$, i.e., $\vvec{x} = \mathcal{R}_{\overline{\vvec{x}}}\left(\vvec{\varepsilon}\right)$, where $\vvec{\varepsilon} \sim \mathcal{N}\left(\vvec{0}_{2\times 1}, \Sigma_\xi\right)$.
We will use this formulation, implemented in the Georgia Tech Smoothing and Mapping (GTSAM) library~\cite{dellaert2012}, to represent and optimize surface normal estimates (detailed in Section \ref{sec:photometric-stereo}).


\section{Related Work}
\vspace{5pt}
Vision-based methods, i.e., SPC, have traditionally been applied to 3D reconstruction for small celestial bodies. 
However, approaches based on active sensors such as Flash-LiDARs have also been proposed. 
Bercovici et al.~\cite{bercovici2019} proposed a pose estimation and shape reconstruction approach based on Flash-LiDAR measurements by solving a maximum likelihood estimation problem via particle-swarm optimization to refine an initial Bezier surface mesh, followed by a least-squares filter providing measurements for the position and orientation of the spacecraft. 
Other works in the field have established proofs-of-concept for batch optimization and graph-based approaches for near-asteroid navigation and shape reconstruction. 
Notably, Nakath et al.~\cite{nakath2020} present an active SLAM framework which also employs Flash-LiDAR as the base measurement, with sensor fusion of data from an inertial measurement unit and star tracker, tested with simulated data. 
However, the limited range of Flash-LiDAR instruments bounds the spacecraft's orbit to unrealistically small radii, reducing the feasible scenarios to either navigation near very small asteroids or the touchdown phase. 
For example, the OSIRIS-REx Guidance, Navigation, and Control (GNC) Flash-LiDAR, which is mentioned in both \cite{nakath2020} and \cite{bercovici2019}, has only a maximum operational range of approximately 1 km and a relatively small $128\times 128$ detector array~\cite{church2020lidar,leonard2022lidar} 

Moreover, the recent OSIRIS-REx mission to Asteroid 101955 Bennu was also equipped with the OSIRIS-REx Laser Altimeter (OLA)~\cite{daly2017} to provide an alternative means of shape reconstruction to SPC. 
The OLA reconstruction process begins by generating local DTMs, similar to SPC, from the range measurements, which are then collated through an iterative closest-point algorithm. 
However, the OSIRIS-REx camera suite (OCAMS)~\cite{rizk2018ocams} features a long-range camera that provides higher spatial resolution than OLA at the same distance. 
Therefore, although the SPC process is more time consuming than the OLA-based approach, SPC products can be available before OLA models, and with a higher resolution. 
Moreover, testing of OLA-generated DTMs showed that uncertainty in OLA measurements created unacceptable errors in the elevations of smaller features and albedo is not automatically included as part of the solution~\cite{lorenz2017}. 
Ultimately, the SPC data products were used in the final touch-and-go phase of the mission. 

More recently, feature-based methods have been shown to be promising technologies for \textit{autonomous} optical navigation and mapping for missions to small bodies. 
Most notably, Dor et al.~\cite{dor2021cvpr} demonstrated precise visual localization and mapping on \textit{real} images of Asteroid 4 Vesta through a feature-based SLAM system based on Oriented FAST and Rotated BRIEF (ORB) features~\cite{rublee2011iccv}. 
ORB is a handcrafted method based on Features from Accelerated Segment Test (FAST) keypoints~\cite{rosten2006eccv} and Binary Robust Independent Elementary Features (BRIEF) descriptors~\cite{calonder2010eccv} and outputs binary descriptor vectors, enabling more efficient matching. 
This work was extended in \cite{dor2022astroslam} to include known dynamical motion constraints between the small body and the spacecraft to further improve mapping and localization performance. 
Furthermore, Driver et al.~\cite{driver2022astrovision} proposed the use of deep learning-based feature detection and description methods, which were shown to significantly outperform traditional handcrafted methods (e.g., ORB), especially in scenarios involving considerable changes in illumination and perspective. 
Our work capitalizes on this recent success in feature-based SLAM and SfM for autonomous optical navigation by imbuing the traditional photogrammetric framework with added characterization power by incorporating photometric stereo constraints for concurrent estimation of surface normals and albedos. 

As stated previously, the proposed approach takes inspiration from the success of SPC. 
However, our implementation has several key differences. 
First, SPC represents surface normals implicitly through a local height map parametrization via the integration of computed surface slopes whereby each DTM is optimized \textit{independently} in a local frame of the associated landmark located at the center of the maplet~\cite{gaskell2008,gaskell2023psj}. 
Conversely, we leverage a minimal representation of the surface normals based on retractions of the tangent space of the $\mathbb{S}^2$ manifold, detailed in Section \ref{sec:unit-vectors}, which allows for global and simultaneous optimization of all observed landmarks, along with their associated surface normals.
Second, and most notably, we forego the dense alignment step used to generate the local DTMs and instead rely on keypoint measurements, which can be generated by autonomous feature detection and description methods, and have been shown to be robust to the significant illumination and perspective changes inherent to small body imagery~\cite{driver2022astrovision}. 
Thus, our framework treats keypoints (referred to as \textit{landmark image locations} in the SPC text~\cite{gaskell2023psj}) as measurements instead of another variable of interest that must be estimated independently. 
Lastly, we exploit the formalism of factor graphs (see Section \ref{sec:factor-graphs}) to model the complex estimation problem, as opposed to the iterative multistep process employed by the traditional SPC implementation~\cite{gaskell2023psj}, allowing for the simultaneous optimization of camera poses, landmark positions, and surface normals and albedos. 

SPC borrows many techniques from the process of \textit{photometric stereo}~\cite{woodham1980}, which has been used extensively in terrestrial applications.
This is not to be confused with \textit{shape-from-shading} (SfS) whereby the shape of a 3D object may be recovered from shading in a \textit{single image}.
However, terrestrial photometric stereo formulations have relied on a number of simplifying assumptions, including Lambertian reflectance~\cite{hayakawa1994photometric,logothetis2019differential} or specialized lighting or image capture setups~\cite{shi2013bi,ikehata2014photometric,cho2018semi}.  
Methods based on deep learning have also been proposed but, as before, require specially designed image acquisition setups~\cite{santo2017dpsn,chen2018psfcn,bi2020deep3d, kaya2023multi}, and thus could not be leveraged in a general multi-view reconstruction scenario. 
We refer the reader to \cite{ackermann2015survey} and \cite{ju2022deep} for more information about physics-based and data-driven approaches, respectively, to photometric stereo for terrestrial applications. 
Also worth mentioning are Neural Radiance Fields (NeRFs)~\cite{mildenhall2021nerf}, which learn a volumetric scene representation via the use of multilayer perceptrons (MLPs) and are primarily applied to view-synthesis. 
However, NeRFs implicitly model the underlying reflectance and material properties of the target object or scene through the learned weights of the MLPs.
Our work, instead, exploits semi-empirical photometric models of the surface of airless bodies with few free parameters to explicitly estimate the topography and material properties of the imaged surface. 


\section{Proposed Approach}
\vspace{5pt}
\begin{figure}[t]
    \centering
    \includegraphics[width=\linewidth]{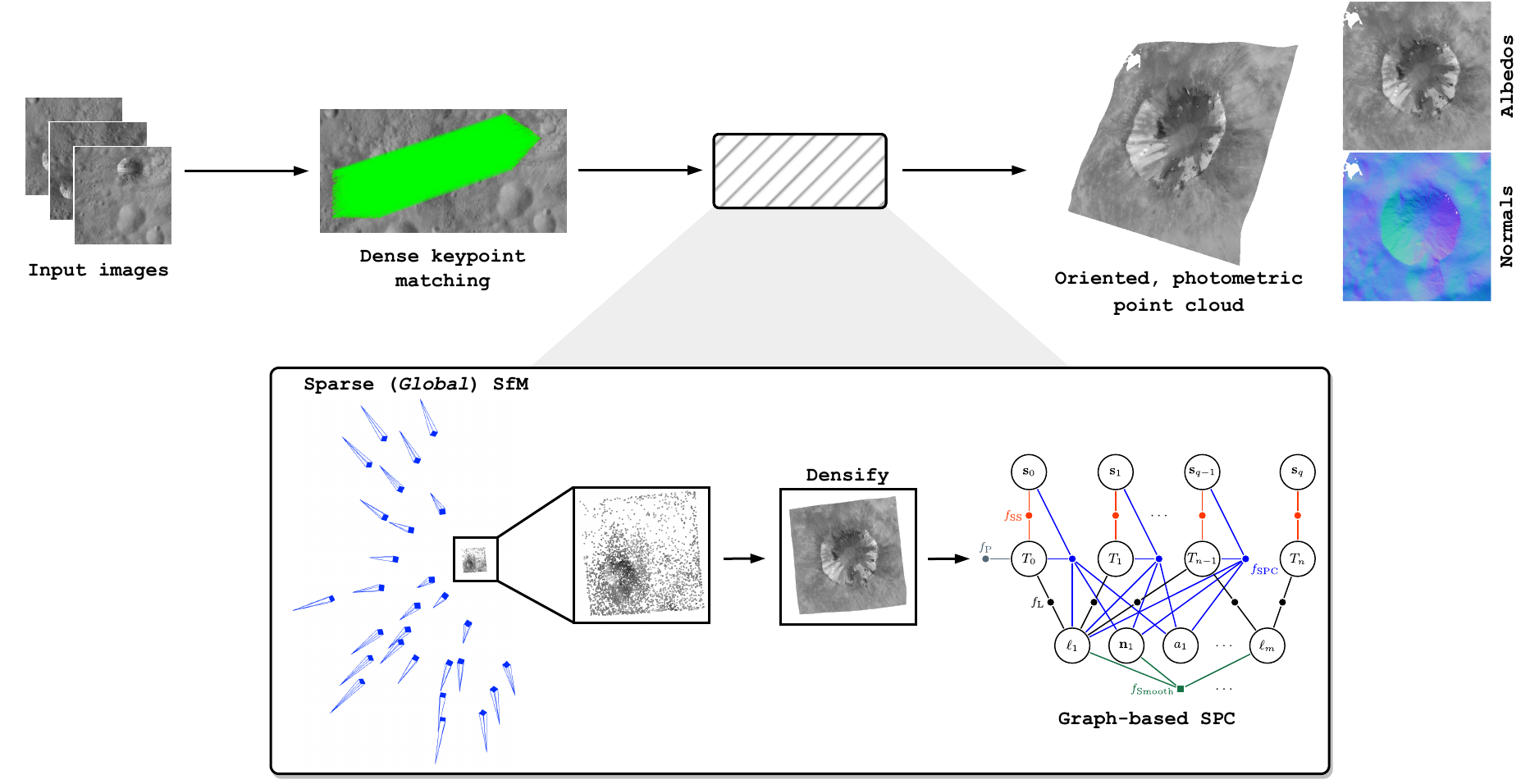}
    \caption{System overview.}
    \label{system-overview}
\end{figure}

First, we introduce the SfM problem (Section \ref{sec:vslam}). 
Next, we formulate SfM as a maximum \textit{a posteriori} (MAP) inference problem using the formalism of \textit{factor graphs} (Section \ref{sec:factor-graphs}). 
Finally, we present the photometric stereo framework and detail its integration into the proposed SPC-SfM system (Sections \ref{sec:photometric-stereo}, \ref{sec:sun-sensor}, and \ref{sec:smoothness}).
An overview of the proposed approach is shown in Figure \ref{system-overview}.


\subsection{Structure-from-Motion} \label{sec:vslam}

The proposed formulation will leverage SfM to estimate the spacecraft's relative pose and a landmark map of the small body's surface.
Feature-based SfM and SLAM~\cite{cadena2016tr} leverage monocular images taken from multiple viewpoints along a trajectory to jointly estimate the robot's pose with respect to its environment and construct a 3D model of the scene.
The SfM architecture is typically comprised of two main components: the \textit{front-end} and the \textit{back-end}.
The front-end extracts 2D interest points (\textit{keypoints}) from images, represents each keypoint with a local feature \textit{descriptor}, and matches keypoints between images by comparing their associated descriptors~\cite{driver2022astrovision}.
The front end also performs \textit{data association} by associating the 2D keypoint measurements with specific points in 3D space (the \textit{landmarks}). 
Finally, the associations from the front-end are used to \emph{simultaneously} reconstruct a map of the environment and resolve the pose of the camera through inference in the \textit{back-end} via maximum \textit{a posteriori} (MAP) estimation. 

Formally, let $\fB$ denote some body-fixed frame of the small body with origin $\oB$, and let $\fC_k$ denote the camera frame at time index $k$ with origin $\oC_k$.
Moreover, let $\vvec{\ell}^\fB_j = \left[\ell_{x,j}^\fB\ \ell_{y,j}^\fB\ \ell_{z,j}^\fB\right]^\top \in \mathbb{R}^3$ denote the vector from $\oB$ to the $j$th surface landmark expressed in $\mathcal{B}$,
let $\vvec{q}^{\fC_k}_j = \left[q_{x,j}^{\fC_k}\ q_{y,j}^{\fC_k}\ q_{z,j}^{\fC_k}\right]^\top \in \mathbb{R}^3$ denote the vector from $\oC_k$ to the $j$th landmark expressed in $\fC_k$, 
and let $\vvec{p}_{j,k} = \left[u_{j,k}\ v_{j,k}\right]^\top \in \mathbb{R}^2$ denote the 2D image coordinates of the $j$th landmark observed by camera $\mathcal{C}_k$, i.e., the keypoint.
SfM seeks the MAP estimate of the camera poses $\mathcal{T} := \left\{T_{\fB\fC_k}\in \mathrm{SE}(3) \mid k = 0,\ldots,n\right\}$ and a network of landmarks (the \textit{map}) $\mathcal{L} := \{\vvec{\ell}^\fB_j \in \mathbb{R}^3 \mid j = 1,\ldots,m\}$ given the (independent) keypoint \textit{measurements} $\mathcal{P} := \{\Hat{\vvec{p}}_{j,k} \in \mathbb{R}^2 \mid k=1,\ldots,n, j=1,\ldots,m\}$:
\begin{align}
    \mathcal{T}^*, \mathcal{L}^* &= \argmax_{\mathcal{T}, \mathcal{L}} p\left(\mathcal{T}, \mathcal{L}\mid \mathcal{P}\right) \\
    &\propto \argmax_{\mathcal{T}, \mathcal{L}} p\left(\mathcal{T}, \mathcal{L}\right) p\left(\mathcal{P} \mid \mathcal{T}, \mathcal{L}\right) \\
    &= \argmax_{\mathcal{T}, \mathcal{L}} p\left(\mathcal{T}, \mathcal{L}\right) \prod_{k} \prod_{j} p\left(\Hat{\vvec{p}}_{j,k} \mid T_{\fB\fC_k}, \vvec{\ell}^\fB_j\right). \label{eq:map}
\end{align}
Note that the SfM solution is innately expressed in some arbitrary body-fixed frame since most SfM techniques assume operation in a static scene, typically referred to as the ``world'' frame~\cite{cadena2016tr}. 
By assuming that the measurements $\Hat{\vvec{p}}_{j,k}$ are corrupted by zero-mean Gaussian noise, i.e., $\Hat{\vvec{p}}_{j,k} = \overline{\vvec{p}}_{j,k} + \vvec{\eta}_{j,k}$ where $\vvec{\eta}_{j,k} \sim \mathcal{N}(\vvec{0},\Sigma_{j,k})$, we get 
\begin{equation}
    p\left(\Hat{\vvec{p}}_{j,k} \mid T_{\fB\fC_k}, \vvec{\ell}^\fB_j\right) \propto \exp\left\{-\frac{1}{2}\|\Pi\left(\vvec{\ell}^\mathcal{B}_{j}, T_{\fB\fC_k}; K\right) - \Hat{\hvec{p}}_{j,k}\|_{\Sigma_{j,k}}^2\right\},
\end{equation}
where $\|\vvec{e}\|^2_{\Sigma} := \vvec{e}^\top\Sigma^{-1}\vvec{e}$, and the \textit{forward-projection} function $\Pi$ relates landmarks $\vvec{\ell}^\mathcal{B}_{j}$ to their (homogenous) coordinates $\hvec{p}_{j,k}$ in the $k$th image, i.e., 
\begin{align} \label{eq:fproj}
    \hvec{p}_{j,k} = \Pi\left(\vvec{\ell}^\fB_{j}, T_{\fB\fC_k}; K\right) &= \frac{1}{d_{j}^{\fC_k}}
    \left[K\,|\,\bm{0}^{3\times1}\right] T_{\fB\fC_k}^{-1} \hvec{\ell}^{\mathcal{B}}_j = \frac{1}{d_{j}^{\fC^k}} K \vvec{q}^{\fC_k}_j,
\end{align}
where $d_{j}^{\fC_k} = q_{z,j}^{\fC_k}$ is the landmark depth in $\fC_k$, $\hvec{\ell}^{\mathcal{B}}_{j} = \left[\left(\vvec{\ell}^\fB_j\right)^\top\ 1\right]^\top \in \mathbb{P}^3$ and $\hvec{p}_{j,k} = \left[\left(\vvec{p}_{j,k}\right)^\top\ 1\right]^\top \in \mathbb{P}^2$ denote the homogeneous coordinates of $\vvec{\ell}^{\fB}_j$ and $\vvec{p}_{j,k}$, respectively, and $K$ is the camera calibration matrix:
\begin{equation}
K = 
    \begin{bmatrix}
    f_x & 0 & c_x \\
    0 & f_y & c_y \\
    0 & 0 & 1
    \end{bmatrix},
\end{equation}
where $f_x$ and $f_y$ are the \textit{focal lengths} in the $x$- and $y$-directions of the camera frame, and $(c_x, c_y)$ is the \textit{principal point} of the camera.

Finally, the MAP estimate can be formulated as the solution to a nonlinear least-squares problem by taking the negative logarithm of \eqref{eq:map}:
\begin{equation} \label{eq:ba}
    \mathcal{T}^*, \mathcal{L}^* = \argmin_{\mathcal{T}, \mathcal{L}} \sum_{k,j} \|\Pi\left(\vvec{\ell}^\fB_j, T_{\fB\fC_k}; K\right) - \Hat{\hvec{p}}_{j,k}\|_{\Sigma_{j,k}}^2,
\end{equation}
where we have omitted the priors $p\left(\mathcal{T}, \mathcal{L}\right)$ for conciseness and generality, which can be ignored if no prior information is assumed (i.e., $p\left(\mathcal{T}, \mathcal{L}\right) = const.$) or can encode relative pose constraints via known dynamical models~\cite{dor2022astroslam}.
This process is commonly referred to as \textit{Bundle Adjustment} (BA). 
Note that the optimization process of SPC decouples estimation of the poses and the landmarks, i.e., \textit{a priori} landmark position and camera pose estimates are passed back-and-forth between the pose determination and DTM construction steps, respectively, until convergence~\cite{gaskell2008}. 

In this work, we will leverage tools from the Georgia Tech Structure-from-Motion (GTSfM) library~\cite{gtsfmcvpr} to generate a sparse SfM solution (described in the following subsection), followed by a densification step to construct a dense map of the imaged surface. 
The keypoint measurements and correspondences, $\Hat{\vvec{p}}_{j,k}$, will be computed by a state-of-the-art, \textit{autonomous} keypoint detection and matching method based on deep learning, i.e., DKM~\cite{edstedt2023dkm}. 
Image brightness measurements will be extracted at observed keypoints and combined with a prior reflectance model (Section \ref{sec:photometric-stereo}) to determine the surface normal and albedo at the estimated landmarks. 
The back-end will leverage factor graphs to represent the MAP estimation problem, which will be discussed in Section \ref{sec:factor-graphs}. 


\begin{figure}[t!]
    \centering
    \includegraphics[width=\linewidth]{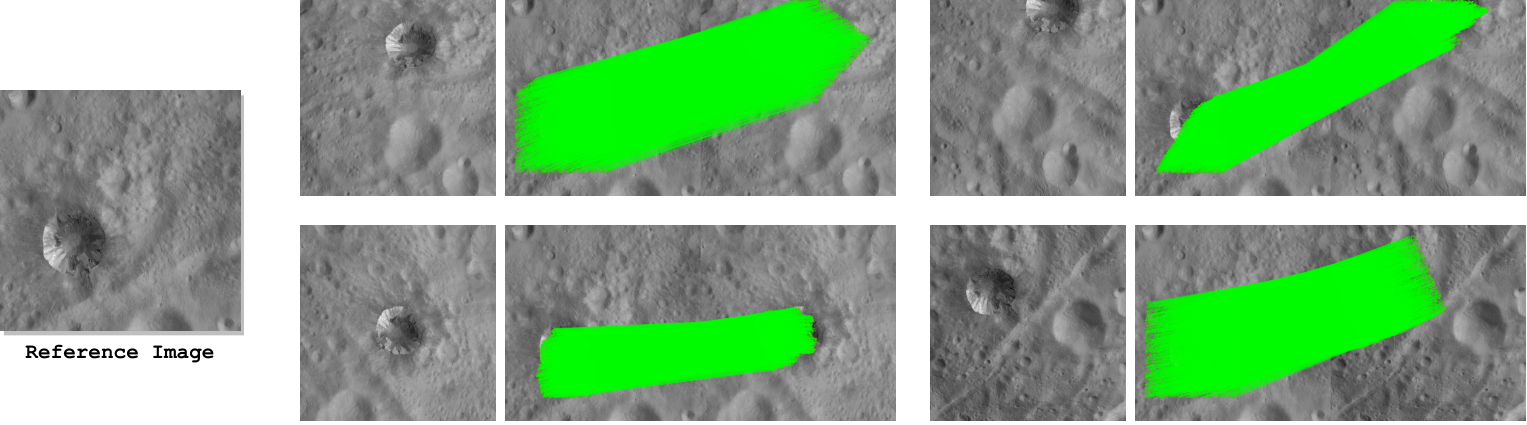}
    \caption{DKM dense matching examples.}
    \label{fig:dkm-matching}
\end{figure}

\subsubsection{Georgia Tech Structure-from-Motion (GTSfM)} \label{sec:gtsfm}

GTSfM~\cite{gtsfmcvpr} is an end-to-end global SfM pipeline based on the Georgia Tech Smoothing and Mapping (GTSAM) library~\cite{dellaert2012}, an estimation toolbox based on factor graphs (see Section \ref{sec:factor-graphs}). 
GTSfM was designed from the ground-up to natively support parallel computation to allow for fast pose estimation and mapping. 
We will leverage GTSfM to generate initial pose estimates and a sparse map, which will then be subsequently densified and used to initialize the graph-based SPC smoothing step (Section \ref{sec:photometric-stereo}). 
Given a collection of images, GTSfM executes the following stages to generate a map (the set of landmarks $\mathcal{L}$) of the imaged target and the pose of the camera when each image was taken  (the poses $\mathcal{T}$):

\paragraph{Correspondence Generation}
Keypoint measurements and correspondences are generated using a state-of-the-art, deep keypoint detection and matching method, Dense Kernalized Feature Matching (DKM)~\cite{edstedt2023dkm}. 
DKM leverages a deep convolutional feature encoder based on ResNet~\cite{he2016deep} to generate coarse features, which are subsequentally matched using a Gaussian Process regression scheme. 
The established coarse matches are then refined at the native resolution of the image using stacked feature maps and large depth-wise kernels to provide dense, \textit{pixel-wise} matches. 
Since these ``detector-free'' methods do not provide a consistent set of keypoints per image, and instead compute a dense mapping between the pixel coordinates of each image pair, we choose a reference image and match all images with respect to this image only.
We provide examples of the dense matching results on \textit{real} images of the Cornelia crater on Asteroid 4 Vesta in Figure \ref{fig:dkm-matching}. 
Only the 2,048 most confident matches are used in the two-view estimation step described next. 

\paragraph{Two-View Estimation} 
The two-view estimation module attempts to estimate the relative rotations, $R_{{\mathcal{C}_l}{\mathcal{C}_k}}$, and translations, $\vvec{r}_{\oC_k\oC_l}^{\fC_l}$, according to the keypoint correspondences from the front-end. 
Specifically, given corresponding keypoints $\vvec{p}_{j,k}$ and $\vvec{p}_{j,l}$ observed by cameras $\fC_k$ and $\fC_l$, respectively, the \textit{essential matrix} $E := [\vvec{r}_{\oC_k\oC_l}^{\fC_l}]_\times R_{\fC_l\fC_k}$ satisfies $\hvec{p}_{j,l}^\top K^{-\top}E K^{-1}\hvec{p}_{j,k} = 0$, where $[\cdot]_\times$ denotes the skew-symmetric cross-product matrix. 
The well-known five-point algorithm~\cite{nister2004} can be used to solve for $E$ given five or more correspondences. 
The essential matrix is robustly estimated using Graph-cut RANSAC~\cite{Barath18cvpr_GCRANSAC}, implemented in OpenCV, with a 0.5 pixel outlier threshold.
Finally, $R_{{\mathcal{C}_l}{\mathcal{C}_k}}$ and $\vvec{r}_{\oC_k\oC_l}^{\fC_l}$ (up to some unknown scale) can be estimated via singular value decomposition (SVD) of the essential matrix and imposing the \textit{cheirality constraint}, i.e., triangulating the keypoints and enforcing that the associated landmark lies in front of the cameras~\cite{forsyth2011modern}. 

\paragraph{Pose Averaging} 
Pose averaging seeks to estimate the ``global'' poses $T_{\fB\fC_k}$ from the previously computed relative poses. 
This module is composed of rotation averaging followed by translation averaging. 
Nominally, GTSfM leverages the Shonan rotation averaging method~\cite{Dellaert20eccv_Shonan}. 
However, since we match all images to a common reference image, this step is trivial as the global orientation of the reference image is taken to be the identity, and the orientations of the remaining images are computed relative to the reference using the previously computed relative orientations. 
Next, given the estimated camera orientations, $R_{\fB\fC_k}$, from the rotation averaging step, along with the relative translations, $\vvec{r}_{\oC_k\oC_l}^{\fC_l}$, from the two-view estimation module, we recover the position, $\vvec{r}^\fB_{\oC_k\oB}$, of each camera (up to an unknown scale) using the method proposed in \cite{Wilson14eccv_1DSfM}. 
We transform the relative translations to a body-fixed frame using the estimated rotations, project them (and landmark rays) along several random directions, and find estimates consistent with the minimum-feedback arc set (MFAS) problem~\cite{Wilson14eccv_1DSfM}, and remove the inconsistent estimates. Finally, we solve for the translations $\mathscr{r} \triangleq \{\vvec{r}^\fB_{\oC_k\oB}\in \mathbb{R}^3 \mid k = 0,\ldots,n\}$ via minimization of a chordal distance $d_\mathrm{ch}(\vvec{u}, \vvec{v}) \triangleq \| \vvec{u}/\|\vvec{u}\| - \vvec{v}/\|\vvec{v}\|\|_2$:
\begin{equation}
    \mathscr{r}^* = \argmin_{\mathscr{r}} \sum_{k,l} d_\mathrm{ch} \left(R_{\fB\fC_l}\vvec{r}_{\oC_k\oC_l}^{\fC_l}, \vvec{r}^\fB_{\oC_k\oB} - \vvec{r}^\fB_{\oC_l\oB} \right)^2,
\end{equation}
where the tuple $(k,l)$ denotes the image indices which were matched in the correspondence generation step.

\paragraph{Data Association}
The keypoint measurements $\Hat{\hvec{p}}_{j,k}$ are \textit{associated} with landmarks from the pairwise matches using a disjoint-set forest (DSF) algorithm. 
Then, the initial 3D position of the landmarks, $\bm{\ell}_j$, are triangulated from the 2D keypoint measurements using the Direct Linear Transform (DLT)~\cite[Chapter~4.1]{hartley2003multiple}, along with RANSAC (with a 0.5 pixel error threshold) to filter possible outlier measurements that made it through the two-view verification process. 

\paragraph{Bundle Adjustment}
Finally, the initial camera poses and triangulated point cloud are refined using a global bundle adjustment step, as described by Equation \ref{eq:ba}, to obtain the final sparse SfM solution.

\paragraph{} The reasoning behind the initial sparse solution is two-fold: (1) leveraging \textit{sparse} correspondences, as opposed to the per-pixel matches, significantly reduces the number of computations in the redundant two-view estimation step; (2) by leveraging only the most confident matches, we reduce the risk of incorporating any outlier matches into the estimation scheme and dense matches can be more reliably verified during the densification step before being added to the map. 
We refer the reader to \cite{gtsfmcvpr} for more details about the GTSfM pipeline. 
Finally, densification of the sparse GTSfM solution is conducted by computing the squared Sampson error~\cite{sampson1982fitting} of each of the putative correspondences and adding the match if its error is below 1. 
The dense map is then triangulated using the same scheme described in the data association module.
This dense map, and the associated camera poses from the sparse solution, are used to initialize the graph-based SPC step described in Section \ref{sec:photometric-stereo}.


\subsection{Factor Graphs} \label{sec:factor-graphs}

\begin{figure}[tb!]
\centering
\begin{subfigure}[t]{\linewidth}
  \centering
  \begin{tikzpicture}[scale=1.0]

    \vertex[label=\color{cadet}\footnotesize$f_{\mathrm{P}}$, draw=cadet, fill=cadet, minimum size=4pt](p) at (-3,0) {};
    
    \vertex[label=center:\small$T_0$, thick, draw=black, minimum size=23pt](T0) at (-2,0) {};
    \vertex[label=center:\small$T_1$, thick, draw=black, minimum size=23pt](T1) at (0,0) {};
    \vertex[label=center:\small$T_{n-1}$, thick, draw=black, minimum size=23pt](Tn1) at (2,0) {};
    \vertex[label=center:\small$T_n$, thick, draw=black, minimum size=23pt](Tn) at (4,0) {};
    
    \vertex[label=center:\small$\bm{\ell}_1$, thick, draw=black, minimum size=23pt](l1) at (-1,-2) {};
    \vertex[label=center:\small$\bm{\ell}_m$, thick, draw=black, minimum size=23pt](lm) at (3,-2) {};
    
    \vertex[label=right:\footnotesize$f_{\mathrm{L}}$,fill=black, minimum size=4pt](m1) at (-1.5,-1) {};
    \vertex[fill=black, minimum size=4pt](m2) at (-.5,-1) {};
    \vertex[fill=black, minimum size=4pt](m3) at (.5,-1) {};
    \vertex[fill=black, minimum size=4pt](m4) at (2.5,-1) {};
    \vertex[fill=black, minimum size=4pt](m5) at (3.5,-1) {};
    
    
    
    % ...
    \vertex[label=center:$\cdots$,draw=none](ee1) at (1,0) {};
    \vertex[label=center:$\cdots$,draw=none](ee2) at (2,-2) {};
    \vertex[draw=none](d2) at (1.2,0) {};
    
    
    \path[draw=cadet, thick, -] (p) edge (T0);
    
    \Edge(T0)(m1)  \Edge(m1)(l1) 
    \Edge(T1)(m2)  \Edge(m2)(l1)
    \Edge(Tn1)(m3) \Edge(m3)(l1)
    \Edge(Tn1)(m4) \Edge(m4)(lm) 
    \Edge(Tn)(m5)  \Edge(m5)(lm)
    

\end{tikzpicture}
  \caption{Typical factor graph formulation of the SfM problem.}
  \label{fig:factor-graph-vslam}
\end{subfigure}\\
\vspace{10pt}
\begin{subfigure}[t]{\linewidth}
  \centering
  \begin{tikzpicture}[scale=1.0]

    \vertex[label=\color{cadet}\footnotesize$f_{\mathrm{P}}$, draw=cadet, fill=cadet, minimum size=4pt](p) at (-3,0) {};
    
    \vertex[label=center:\small$T_0$, thick, draw=black, minimum size=23pt](T0) at (-2,0) {};
    \vertex[label=center:\small$T_1$, thick, draw=black, minimum size=23pt](T1) at (0,0) {};
    \vertex[label=center:\small$T_{n-1}$, thick, draw=black, minimum size=23pt](Tn1) at (2,0) {};
    \vertex[label=center:\small$T_n$, thick, draw=black, minimum size=23pt](Tn) at (4.2,0) {};
    
    \vertex[label=center:\small$\bm{\ell}_1$, thick, draw=black, minimum size=23pt](l1) at (-1,-2) {};
    \vertex[label=center:\small$\bm{\ell}_m$, thick, draw=black, minimum size=23pt](lm) at (3.5,-2) {};
    
    \vertex[label=center:\small$\mathbf{n}_1$, thick, draw=black, minimum size=23pt](n1) at (.25,-2) {};
    
    \vertex[label=center:\small$a_1$, thick, draw=black, minimum size=23pt](a1) at (1.5,-2) {};
    
    \vertex[label=center:\small$\mathbf{s}_0$, thick, draw=black, minimum size=23pt](s0) at (-2,2) {};
    \vertex[label=center:\small$\mathbf{s}_1$, thick, draw=black, minimum size=23pt](s1) at (0,2) {};
    \vertex[label=center:\small$\mathbf{s}_{q-1}$, thick, draw=black, minimum size=23pt](sk1) at (2,2) {};
    \vertex[label=center:\small$\mathbf{s}_q$, thick, draw=black, minimum size=23pt](sk) at (4.2,2) {};
    
    \vertex[draw = blue, fill=blue, minimum size=4pt](fps0) at (-1,0) {};
    \vertex[draw = blue, fill=blue, minimum size=4pt](fps1) at ( 1,0) {};
    \vertex[label=south east:\color{blue}\footnotesize$f_{\mathrm{SPC}}$, draw = blue, fill=blue, minimum size=4pt](fps2) at ( 3,0) {};
    
    \path[draw=blue, thick, -] (s0) edge (fps0);
    \path[draw=blue, thick, -] (T0) edge (fps0);
    \path[draw=blue, thick, -] (l1) edge (fps0);
    \path[draw=blue, thick, -] (n1) edge (fps0);
    \path[draw=blue, thick, -] (a1) edge (fps0);
    
    \path[draw=blue, thick, -] (s1) edge (fps1);
    \path[draw=blue, thick, -] (T1) edge (fps1);
    \path[draw=blue, thick, -] (l1) edge (fps1);
    \path[draw=blue, thick, -] (n1) edge (fps1);
    \path[draw=blue, thick, -] (a1) edge (fps1);
    
    \path[draw=blue, thick, -] (sk1) edge (fps2);
    \path[draw=blue, thick, -] (Tn1) edge (fps2);
    \path[draw=blue, thick, -] (l1) edge (fps2);
    \path[draw=blue, thick, -] (n1) edge (fps2);
    \path[draw=blue, thick, -] (a1) edge (fps2);
    
    \vertex[label=left:\footnotesize$f_{\mathrm{L}}$,fill=black, minimum size=4pt](m1) at (-1.5,-1) {};
    \vertex[fill=black, minimum size=4pt](m2) at (-.5,-1) {};
    \vertex[fill=black, minimum size=4pt](m3) at (.5,-1) {};
    \vertex[fill=black, minimum size=4pt](m4) at (2.75,-1) {};
    \vertex[fill=black, minimum size=4pt](m5) at (3.85,-1) {};
    
    \vertex[label=left:\color{coquelicot}\footnotesize$f_{\mathrm{SS}}$, draw=coquelicot, fill=coquelicot, minimum size=4pt](fss0)  at (-2,1) {};
    \vertex[draw=coquelicot, fill=coquelicot, minimum size=4pt](fss1)  at ( 0,1) {};
    \vertex[draw=coquelicot, fill=coquelicot, minimum size=4pt](fssk1) at ( 2,1) {};
    \vertex[draw=coquelicot, fill=coquelicot, minimum size=4pt](fssk)  at ( 4.2,1) {};

    \node[rectangle, label=left:\color{darkspringgreen}\footnotesize$f_{\mathrm{Smooth}}$, draw=darkspringgreen, fill=darkspringgreen, minimum size=4pt](fsm0) at (1.5,-3) {};
    
    
    
    % ...
    \vertex[label=center:$\ldots$,draw=none](ee1) at (1,1) {};
    \vertex[label=center:$\ldots$,draw=none](ee2) at (2.5,-2) {};
    \vertex[label=center:$\ldots$,draw=none](ee2) at (2.5,-3) {};
    \vertex[draw=none](d2) at (1.2,0) {};
    
    \path[draw=cadet, thick, -] (p) edge (T0);
    
    \Edge(T0)(m1) \Edge(m1)(l1) 
    \Edge(T1)(m2) \Edge(m2)(l1)
    \Edge(Tn1)(m3) \Edge(m3)(l1)
    \Edge(Tn1)(m4) \Edge(m4)(lm) 
    \Edge(Tn)(m5)  \Edge(m5)(lm)
    
    \path[draw=coquelicot, thick, -] (s0) edge (fss0);
    \path[draw=coquelicot, thick, -] (T0) edge (fss0);
    \path[draw=coquelicot, thick, -] (s1) edge (fss1);
    \path[draw=coquelicot, thick, -] (T1) edge (fss1);
    \path[draw=coquelicot, thick, -] (sk1) edge (fssk1);
    \path[draw=coquelicot, thick, -] (Tn1) edge (fssk1);
    \path[draw=coquelicot, thick, -] (sk) edge (fssk);
    \path[draw=coquelicot, thick, -] (Tn) edge (fssk);

    \path[draw=darkspringgreen, thick, -] (l1) edge (fsm0);
    \path[draw=darkspringgreen, thick, -] (lm) edge (fsm0);
    \path[draw=darkspringgreen, thick, -] (n1) edge (fsm0);

\end{tikzpicture}
  \caption{With proposed factors \textcolor{blue}{$f_{\mathrm{SPC}}$} relating to photometric stereo constraints, \textcolor{coquelicot}{$f_{\mathrm{SS}}$} relating to Sun sensor measurements, and \textcolor{darkspringgreen}{$f_{\mathrm{Smooth}}$} relating to local smoothness constraints.}
  \label{fig:factor-graph-ps}
\end{subfigure}%
\caption{Variable nodes are camera poses $T_k \in \mathrm{SE}(3)$, landmarks $\bm{\ell}_j \in \mathbb{R}^3$, sun vectors $\mathbf{s}_l \in \mathbb{S}^2$, surface normals $\mathbf{n}_i \in \mathbb{S}^2$, and surface albedos $a_i \in [0, 1]$. Factor nodes $f_\mathrm{L}$ and \textcolor{cadet}{$f_\mathrm{P}$} relate to keypoint-based landmark measurements and possibly a prior factor, respectively.}
\label{fig:factor-graph}
\end{figure}
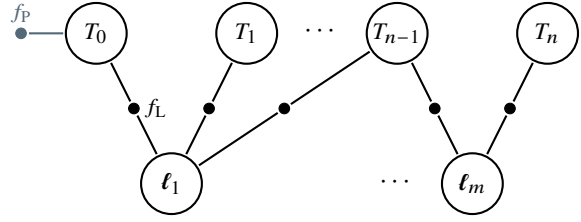
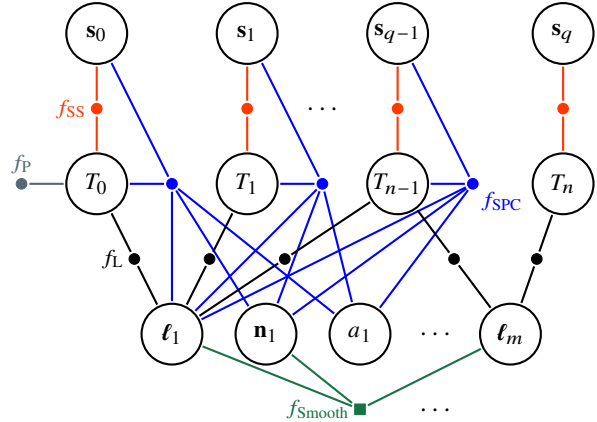

We leverage the formalism of factor graphs to facilitate the fusion of measurements from onboard sensors (i.e., monocular cameras and Sun sensors) through prior photometric models to \textit{simultaneously} estimate the camera poses, landmark positions, and surface normals and albedos at each landmark. 
Formally, a factor graph is a bipartite graph $G = \left(\mathcal{F}, \Theta, \mathcal{E}\right)$ with \textit{factor nodes} $f_i \in \mathcal{F}$ that abstract the measurements and prior knowledge $z_i \in \mathcal{Z}$ as generalized probabilistic constraints between \textit{variable nodes} $\theta_j \in \Theta$, the unknown random variables, where \textit{edges} $e_{i,j} \in \mathcal{E}$ define the interdependence relationships between a factor $f_i$ and a variable $\theta_j$.
With these definitions, a factor graph $G$ defines a factorized function
\begin{equation} \label{eq:f}
    f(\Theta) = \prod_i f_i\left(\Theta_i\right),
\end{equation}
where each measurement factor $f_i(\Theta_i) = l(\Theta_i; z_i)$ of the variables $\Theta_i = \{\theta_j \in \Theta \given e_{i,j} \in \mathcal{E}\}$, with likelihood $l\left(\Theta_i; z_i\right) \propto p\left(z_i \given \Theta_i\right)$, and each prior factor $f(\Theta_i) = p(\Theta_i)$ represents a term in the joint probability density function (PDF), i.e., $f(\Theta) \propto p\left(\Theta \given \mathcal{Z}\right)$. 
We seek the variable assignment $\Theta^*$ through maximum \textit{a-posteriori} (MAP) inference over the joint probability distribution encoded by the factors in the factor graph: 
\begin{equation} \label{eq:Theta_opt}
    \Theta^* = \argmax_\Theta \prod_i f_i\left(\Theta_i\right).
\end{equation}
Assuming a zero-mean Gaussian noise model with measurement covariance $\Sigma_i$ gives factors of the form 
\begin{equation} \label{eq:fi}
    f_i(\Theta_i) \propto \exp\left\{-\frac{1}{2} \|h_i(\Theta_i) - z_i\|^2_{\Sigma_i}\right\},
\end{equation}
where $h_i(\cdot)$ is a measurement prediction function. 
Moreover, assume that priors $p(\Theta_i)$ can be written as $p(\Theta_i) \propto \exp\left\{-\frac{1}{2} \|h_i(\Theta_i) - z_i\|^2_{\Sigma_i}\right\}$ with prior mean and covariance $z_i$ and $\Sigma_i$, respectively. 
Therefore, solving \eqref{eq:Theta_opt} is equivalent to minimizing a sum of nonlinear least-squares via
\begin{equation} \label{eq:logf}
    \argmin_\Theta(-\log f(\Theta)) = \argmin_\Theta \frac{1}{2}\sum_i \|h_i(\Theta_i) - z_i\|^2_{\Sigma_i}.
\end{equation}
This formulation allows factor graphs to support PDFs or cost functions of any number of variables~\cite{kaess2012}, allowing for the inclusion of multiple sensor modalities, as well as prior knowledge and constraints to uniquely determine the MAP solution for the unknown variables $\Theta^*$. 
The typical factor graph formulation of SfM is shown in Fig. \ref{fig:factor-graph-vslam}, where the factors $f_\mathrm{L}\left(\vvec{\ell}^\fB_k, T_{\fB\fC_k}; K\right)$ relate to the forward-projection error function defined in Equation \eqref{eq:ba}. 

Solving the nonlinear least-squares problem in \eqref{eq:logf} typically involves repeated linearization. 
For nonlinear measurement functions $h_i(\cdot)$, nonlinear optimization approaches such as the Levenberg-Marquardt algorithm (LMA) leverage repeated first-order linear approximations to \eqref{eq:logf} to approach the minimum. 
In addition, the interdependence relationships encoded by the edges of the factor graph capture the factored nature of the PDF and sparsity of the underlying information matrix, allowing for \textit{exact} nonlinear optimization in an \textit{incremental} setting by exploiting the sparse edge connections to identify the variables to be optimized when a new measurement becomes available~\cite{kaess2012}. 
The factor graph formulation of the proposed keypoint-based SPC problem is implemented using the Georgia Tech Smoothing and Mapping (GTSAM) library~\cite{dellaert2012}, an estimation toolbox based on factor graphs pioneered at Georgia Tech. 
The toolbox provides a fully customizable framework for factor graph construction and a suite of nonlinear optimization methods. 

\subsection{Small Body Photometry} \label{sec:photometry}

\tdplotsetmaincoords{70}{120}
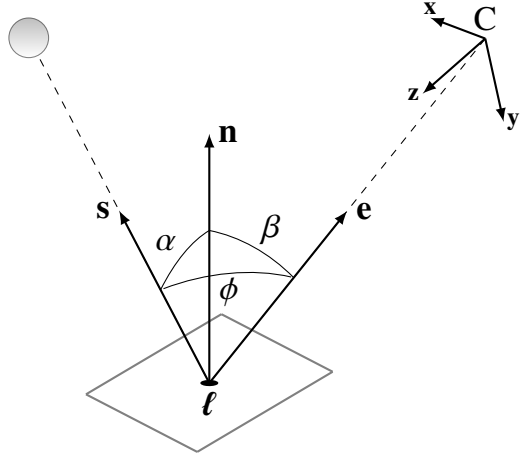
\begin{figure}[t]
\centering




\tdplotsetmaincoords{70}{120}
\begin{tikzpicture}[tdplot_main_coords, scale=1.2, every node/.style={scale=1.3}]

    \coordinate (oo) at (0,0,0);
    \coordinate (bl) at (-4/2,-2/2,0);
    \coordinate (br) at (2/2,-2/2,0);
    \coordinate (tl) at (-2/2,2/2,0);
    \coordinate (tr) at (4/2,2/2,0);
    \coordinate (ss) at (3/0.75, 0, 4/0.75);
    \coordinate (ss2) at (3/0.8, 0, 4/0.8);
    \coordinate (cc) at (0, 3/.85, 4/.85);
    \coordinate (cx) at (0/.85, 2.4/.85, 4.1/.85);
    \coordinate (cy) at (0/.85, 3.2/.85, 3.2/.85);
    \coordinate (cz) at (0, 3/1.1, 4.3/1.1);
    
    \draw[thick,gray] (bl) -- (br);
    \draw[thick,gray] (bl) -- (tl);
    \draw[thick,gray] (tl) -- (tr);
    \draw[thick,gray] (tr) -- (br);
    
    \node[inner sep=0pt] (sun) at (3/0.95, 0, 4/0.95) {};
    \shadedraw[opacity = .5] (ss) circle (.22cm);
    
    \node[inner sep=0pt] (camera) at (0, 3/1.1, 4/1.1) {};
    \node[inner sep=0pt, label=above:{$\mathrm{C}$}] (co) at (0, 3/.85, 4/.85) {};
    \draw[thick,->] (cc) -- (cz) node [black,left] {\scriptsize{$\mathbf{z}$}};
    \draw[thick,->] (cc) -- (cx) node [black,above] {\scriptsize{$\mathbf{x}$}};
    \draw[thick,->] (cc) -- (cy) node [black,right] {\scriptsize{$\mathbf{y}$}};
    

    \draw[dashed] (ss2) -- (oo);
    \draw[dashed] (cc) -- (oo);
    \node[label=below:{$\bm{\ell}$}] (a) at (0,0,0) {};
    \fill[black] (oo) circle (0.10);
    \draw[thick,->] (oo) -- (0,0,5/1.7) coordinate (nhat) node [black,right] {$\,\mathbf{n}$};
    \draw[thick,->] (oo) -- (3/1.5,0,4/1.5) coordinate (ihat) node [black,left] {$\mathbf{s}\,$};
    \draw[thick,->] (oo) -- (0,3/1.7,4/1.7) coordinate (ehat) node [black,right] {$\,\mathbf{e}$};
    
    \tdplotsetrotatedcoords{0}{-90}{0}
    \tdplotdrawarc[tdplot_rotated_coords]{(0,0,0)}{1.8}{0}{90-53.13}{anchor=south}{$\quad\beta$}
    \tdplotsetrotatedcoords{90}{-90}{0}
    \tdplotdrawarc[tdplot_rotated_coords]{(0,0,0)}{1.8}{0}{-36.87}{anchor=south}{$\alpha\quad$}
    \tdplotsetrotatedcoords{45}{-62.0616}{0}
    \tdplotdrawarc[tdplot_rotated_coords]{(0,0,0)}{1.8}{24}{-24}{anchor=north}{$\phi$}
\end{tikzpicture}

\caption{Photometry conventions.}
\label{fig:angles}
\end{figure}

\begin{table}[]
    \centering
    \ra{2.5}
    \caption{Investigated reflectance functions. The coefficients for the Schr\"{o}der model are listed in Table \ref{tab:sch-coeff}.}
    \begin{tabular}{lcccc}
        \toprule
        & $r_F(\alpha, \beta, \phi)$ & $g(\phi)$ & $\Lambda(\phi)$ & \\
        \midrule
        \rowcolor[gray]{0.9}
        Lunar-Lambert & $a\left((1 - g(\phi))\cos\alpha + g(\phi)\frac{2\cos\alpha}{\cos\alpha + \cos\beta}\right)$ & $\exp(-\phi/60)$ & --- & \parbox{1cm}{\begin{equation}\label{eq:ref-ll}\end{equation}} \\
        Schr\"{o}der & $a\Lambda(\phi)\left((1 - g(\phi))\cos\alpha + g(\phi)\frac{2\cos\alpha}{\cos\alpha + \cos\beta}\right)$ & $C_0 + C_1\phi$ & $\sum_{i=0}^4 B_i\phi^i$ & \parbox{1cm}{\begin{equation}\label{eq:ref-sch}\end{equation}} \\
        \bottomrule
    \end{tabular}
    \label{tab:rad-factors}
\end{table}

\begin{table}[]
    \centering
    \ra{1.0}
    \caption{Coefficients for the Schr\"{o}der reflectance function~\cite{schroder2013resolved}, where we have normalized $B_i$, $i=1,2,3,4$, such that $\Lambda(0) = 1$.}
    \begin{tabular}{cccccccc}
        \toprule
        $C_0$ & $C_1$ && $B_0$ & $B_1$ & $B_2$ & $B_3$ & $B_4$ \\
        \midrule
        $0.830$ & $-7.22\times 10^{-3}$ &&  $1$ & $-1.7160\times 10^{-2}$ & $1.8306\times 10^{-4}$ & $-1.0399\times 10^{-6}$ & $2.3223\times 10^{-9}$ \\
        \bottomrule
    \end{tabular}
    \label{tab:sch-coeff}
\end{table}


\textit{Photogrammetry}, serving as the theoretical foundation for contemporary techniques such as SfM, primarily concentrates on establishing geometric relationships between points in an image (the \textit{keypoints}) and the corresponding points in the scene (the \textit{landmarks}). In contrast, \textit{photometry} attempts to determine the ``brightness'' of a point within the scene as it appears in the observed image.
Photogrammetry has historically garnered more attention in the computer vision community, due, in part, to the relatively low complexity as compared to photometry.
Indeed, photometric modeling is inherently complex. 
This is especially true for terrestrial applications, which typically require the use of deep neural networks to achieve accurate photometric reconstructions~\cite{mildenhall2020nerf}. 
However, operation in space presents us with a number of advantages that simplify the photometric modeling process. 
First, we may treat the Sun as a point source delivering collimated light to the surface owing to the large distance to the Sun (i.e., the Sun subtends an angle of $0.5^\circ$ at Earth~\cite{shepard2017}) and the lack of atmosphere of most small bodies to scatter the incoming light. 
Second, the direction of the incoming light can be precisely measured using typical onboard instrumentation (e.g., Sun sensors, star trackers). 
Third, previous photometric modeling of small bodies has demonstrated that \textit{global} reflectance functions, as opposed to \textit{spatially-varying} reflectance functions, can precisely estimate the observed brightness across the surface of a target small body.
Moreover, these global reflectance properties are similar across asteroids of the same taxonomic class~\cite{li2015}.

More formally, an image may be considered as a mapping $I \colon \Omega \rightarrow \mathbb{R}_+$ over the pixel domain $\Omega \subset \mathbb{R}^2$ that maps points in the image, $\vvec{p} \in \Omega$, to their corresponding ``brightness" value, $I(\vvec{p}) \in \mathbb{R}_+$. 
Here, ``brightness'' refers to the fact that the image values correspond to the amount of light falling on the photodetector inside the camera, referred to as the image \textit{irradiance} with units of power per unit area (W$\cdot$m$^{-2}$). 
It can be shown~\cite[Chapter~10.3]{horn1986} that the image irradiance is proportional to the \textit{radiance}, with units of power per unit solid angle per unit area (W$\cdot$sr$^{-1}$$\cdot$m$^{-2}$), reflected towards the camera from the surface. 
Thus, each point $\vvec{p} \in \Omega$ corresponds to the radiance emitted from a point (or, more precisely, an infinitesimal patch) on the surface of the body. 
As such, the raw image brightness values, digitized to a 14-bit ``digital number'' (DN) for the Dawn Framing Camera~\cite{sierks2011dawn}, may be converted to radiance by normalizing by the image exposure time and utilizing an appropriate \textit{radiometric calibration} process~\cite{schroder2013cal,schroder2014cal}.

The emitted radiance from the surface, $L(\alpha, \beta, \phi, a)$ and the incident (collimated) irradiance from the Sun, $F$, which is a function of the distance from the target body to the Sun~\cite{shepard2017}, are related by the \textit{bidirectional reflectance function}, $r$ (in units of $\text{sr}^{-1}$):
\begin{equation}
    \frac{L(\alpha, \beta, \phi, a)}{F} = r(\alpha, \beta, \phi, a),
\end{equation}
where $a$ is the surface albedo, $\alpha$ is the angle between the incoming light and the surface normal, or the \textit{incidence angle}, $\beta$ is the angle between the emitted light and the surface normal, or the \textit{emission angle}, and $\phi$ is the angle between the emitted light and the incoming light, or the \textit{phase angle} (see Figure \ref{fig:angles}). 
A similar measure of reflectance, which is very popular in the context of planetary photometry~\cite{shepard2017, li2015}, is the \textit{bidirectional radiance factor}, $r_F$ (in units of $I/F$, which are dimensionless), which is the ratio of the bidirectional reflectance of a surface to that of a perfect Lambertian surface illuminated and viewed from overhead (i.e., $\alpha = \beta = \phi = 0$):
\begin{equation} \label{eq:rad-factor}
    r_F(\alpha, \beta, \phi, a) = \pi r(\alpha, \beta, \phi, a). 
\end{equation}
Henceforth, we are referring to the radiance factor when mentioning the reflectance function. 

We consider two reflectance functions in this investigation, defined in Table \ref{tab:rad-factors}.
The first model, which we refer to as the \textit{Lunar-Lambert} function, defined in Equation \eqref{eq:ref-ll}, features a combination of Lambert and Lommel-Seeliger photometric functions weighted according to an exponential function of the phase angle, $g(\phi)$. 
The model in Equation \eqref{eq:ref-ll} is an approximation to McEwen's reflectance function~\cite{mcewen1991, mcewen1996precise}, which was fit to lunar imagery captured by the Galileo spacecraft. 
This model was chosen because it is nominally used by SPC and has been shown to be well-suited for photometry on a wide range of small bodies~\cite{gaskell2008, raymond2011, barnouin2020, gaskell2023psj}.
The second, which we refer to as the \textit{Schr\"{o}der} model~\cite{schroder2013resolved}, defined in Equation \eqref{eq:ref-sch}, again features a combination of Lambert and Lommel-Seeliger photometric functions, but is weighted according to an \textit{affine} function of the phase angle and also features a \textit{surface phase function}, $\Lambda(\phi)$. 
The \textit{phase function} describes the change in brightness with phase that are independent of the incidence and emission angles. 
The Schr\"{o}der model in Equation \eqref{eq:ref-sch} was fit to approach imagery of Asteroid 4 Vesta captured during the Dawn mission. 
By investigating this reflectance model, we will demonstrate the utility of specialized reflectance models fit to relevant imagery. 

We will discuss how these photometric principles can be incorporated into a feature-based SfM system in the following section.
We will consider both calibrated and uncalibrated imagery in this investigation. 
For the uncalibrated case, we use corrected DN values where various error sources such as dark current and readout smear have been removed~\cite{schroder2013cal,schroder2014cal}. 
For the calibrated case, the images have been converted to units of $I/F$ according to the radiometric calibration process detailed in \cite{schroder2014cal}.


\subsection{Photometric Stereo Constraints} \label{sec:photometric-stereo}

Photometric stereo techniques are integrated into the feature-based SfM system to estimate surface normals and albedos at estimated landmarks. 
Photometric stereo~\cite{woodham1980} is the process of determining surface gradients of an object by observing it from different viewpoints and lighting conditions and is leveraged by SPC to facilitate dense surface reconstruction. 
As before, let an image taken at time index $k$ be denoted by $I_k \colon \Omega \rightarrow \mathbb{R}_+$ over the pixel domain $\Omega \subset \mathbb{R}^2$.
The measured image brightness $\Hat{I}_k(\vvec{p}_{j,k})$ (calibrated to units of $I/F$) at a keypoint $\vvec{p}_{j,k} \in \Omega$ in image $I_k$ associated with a landmark $\bm{\ell}_j \in \mathbb{R}^3$ can be modeled by an appropriate reflectance function, as detailed in Section \ref{sec:photometry}:
\begin{equation} \label{eq:Ik_angles}
    I(\alpha_{j,k}, \beta_{j,k}, \phi_{j,k}, a_j) = r_F(\alpha_{j,k}, \beta_{j,k}, \phi_{j,k}, a_j),
\end{equation}
where $a_j$ is the albedo at landmark $\vvec{\ell}_j$ and $\alpha_{j,k}$, $\beta_{j,k}$, and $\phi_{j,k}$  are the incidence, emission, and phase angles, respectively, at landmark $\vvec{\ell}_j$ in the $k^{th}$ image.
When considering \textit{uncalibrated} imagery, a scale, $\lambda_k$, and bias, $\xi_k$, term are typically included in Equation \eqref{eq:rad-factor} to account for factors such as distance to the Sun, exposure times, and background noise for each image~\cite{gaskell2008,gaskell2023psj,alexandrov2018multiview}, i.e.,
\begin{equation}
    I(\alpha_{j,k}, \beta_{j,k}, \phi_{j,k}, a_j) = \lambda_k r_F(\alpha_{j,k}, \beta_{j,k}, \phi_{j,k}, a_j) + \xi_k. 
\end{equation}
In this case, $a_j$ is the \textit{relative} surface albedo, which refers to the fact that, unless considering radiometrically calibrated imagery~\cite{schroder2013cal,schroder2014cal}, the albedo $a_j$ is only proportional to the absolute albedo.

The Sun-relative direction $\mathbf{s}^\fB_{k} \in \mathbb{S}^2$ in $I_k$, expressed in the body-fixed frame of the small body $\mathcal{B}$, can be estimated using measurements from typical onboard instrumentation (e.g., Sun sensors, star trackers), detailed in Section \ref{sec:sun-sensor}.  
The emitted light vector $\mathbf{e}^\fB_{k,j} = \vvec{r}^\fB_{\oC_k\oB} - \vvec{\ell}^\fB_j$ can be determined from the estimates of $T_{\fB\fC_k}$ and $\vvec{\ell}^\fB_j$ provided by SfM.  
Finally, dropping the superscripts and letting $T_k$ denote $T_{\fB\fC_k}$ for conciseness, Equation \eqref{eq:Ik_angles} can be written in terms of $\vvec{s}_k$, $\vvec{e}_{k,j}$, and the surface normal $\vvec{n}_j \in \mathbb{S}^2$ at $\vvec{\ell}_j$ (see Fig. \ref{fig:angles}) by noticing that  $\cos\alpha_{j,k} = \mathbf{s}_k^\top\mathbf{n}_j$, $\cos\beta_{j,k} = \mathbf{e}_{k,j}^\top\mathbf{n}_j/\|\mathbf{e}_{k,j}\|$, and $\phi_{j,k} = \cos^{-1}\left(\mathbf{s}_k^\top\mathbf{e}_{k,j}/\|\mathbf{e}_{k,j}\|\right)$. 
For example, the Lunar-Lambert model (Equation \eqref{eq:ref-ll}) becomes
\begin{equation} \label{eq:Ik_vecs}
    I(T_k, \mathbf{s}_k, \bm{\ell}_j, \mathbf{n}_j, a_j) = a_j\left((1 - g(\mathbf{s}_k, \mathbf{e}_{k,j}))\mathbf{s}_k^\top\mathbf{n}_j + \vphantom{\frac{\mathbf{1}^\top_k}{\|\mathbf{1}^\top_k\|}} g(\mathbf{s}_k, \mathbf{e}_{k,j})\frac{2\mathbf{s}_k^\top\mathbf{n}_j}{\mathbf{s}_k^\top\mathbf{n}_j + \mathbf{e}_{k,j}^\top\mathbf{n}_j/\|\mathbf{e}_{k,j}\|}\right).
\end{equation}
We can now define a factor $f_{\mathrm{SPC}}$ corresponding to the presented photometric stereo constraints (assuming zero-mean Gaussian noise) as follows:
\begin{equation} \label{eq:fSPC}
    f_{\mathrm{SPC}}(T_k, \mathbf{s}_k, \vvec{\ell}_j, \vvec{n}_j, a_j; \Sigma_k) \propto \exp\left\{-\frac{1}{2} |I(T_k, \mathbf{s}_k, \vvec{\ell}_j, \vvec{n}_j, a_j) - \Hat{I}_k(\Hat{\vvec{p}}_{j,k})|^2_{\Sigma_k}\right\}.
\end{equation}
This allows for estimation of $\vvec{n}_j$ and $a_j$ using the measurements $\Hat{I}_k(\Hat{\vvec{p}}_{j,k})$, while also further constraining the landmark's position $\vvec{\ell}_j$, Sun-relative direction $\vvec{s}_k$, and the position of the spacecraft $\vvec{r}_{\oC_k\oB}$. 
The corresponding factor graph diagram is shown in Fig. \ref{fig:factor-graph-ps}.



\subsection{Sun Vector Measurements} \label{sec:sun-sensor}

Our framework assumes knowledge of the Sun-relative direction $\vvec{s}_k$. 
This direction is usually determined from the navigation solution and ephemeris files, and onboard attitude estimates (e.g., from a star tracker) allow this direction to be expressed in the camera frame. 
The Sun direction may also be measured directly by a Sun sensor. 
For example, the OSIRIS-REx spacecraft featured multiple coarse Sun sensors, each with an accuracy of $\pm 1^{\circ}\ (3\sigma)$~\cite{bierhaus2018}. Moreover, fine Sun sensors have accuracy on the order of $\pm 0.01^{\circ}\ (3\sigma)$. 
For simplicity, we will assume that the measurements $\Hat{\vvec{s}}^\fC_k  \in \mathbb{S}^2$ are available at each time index $k$ and are expressed in the camera frame $\fC$ via the fixed relative pose $T_{\mathcal{U}\fC}$ of the Sun sensor frame $\mathcal{U}$ with respect to $\fC$. 
Recalling that $T_k$ denotes $T_{\fB \fC_k}$ and $\vvec{s}_k$ is expressed in the $\fB$ frame, a measurement prediction function $\vvec{s}^C(T_k, \vvec{s}_k)$ can be defined to predict the measured incident light direction $\hat{\vvec{s}}^\fC$ in the $\fC$ frame from the current estimates of $T_k$ and $\vvec{s}_k$:
\begin{equation}
    \mathbf{s}^\mathcal{C}(T_{k}, \mathbf{s}_k) \triangleq R_{\fB\fC_k}^{-1}\mathbf{s}_k.
\end{equation}
We define a factor $f_{SS}$ to incorporate (simulated) Sun sensor measurements into the estimation problem as follows:
\begin{equation} \label{eq:fSS}
    f_{\mathrm{SS}}\left(T_k, \mathbf{s}_k; \Sigma_{j,k}\right) \propto \exp\left\{-\frac{1}{2}\|\mathbf{s}^\mathcal{C}(T_{k}, \mathbf{s}_k) - \Hat{\mathbf{s}}^\mathcal{C}_k\|_{\Sigma_{j,k}}^2\right\}.
\end{equation}
This further constrains the orientation of the camera $R_{\fB\fC_k}$ and the Sun-relative direction $\vvec{s}_k$.


\subsection{Local Smoothness Constraints} \label{sec:smoothness}

While the photometric minimization and Sun vector terms modeled by $f_\mathrm{SPC}$ and $f_\mathrm{SS}$, respectively, are sufficient to estimate the surface normal and albedo, Horn~\cite{horn1990} indicates that the solution tends to be unstable and gets stuck in local minima, especially if starting far from the solution. 
This has also been demonstrated in other works on small body shape reconstruction~\cite{capanna2013spc,alexandrov2018multiview}.
Thus, Horn proposed the use of local smoothness constraints which minimize the ``departure from smoothness.'' 
Our smoothness constraint \textit{factors} are defined as follows:
\begin{equation} \label{eq:fsmooth}
    f_\mathrm{Smooth}\left(\vvec{\ell}_j, \mathbf{n}_j, \vvec{\ell}_{j'}; \eta\right) \propto \exp\left\{-\frac{1}{2}\eta\left|\cos^{-1}\left(\mathbf{d}_{j',j}^\top\mathbf{n}_j\right) - 90^\circ\right|^2\right\},
\end{equation}
where $\eta$ weights the local smoothness penalty, and $\vvec{d}_{j',j} = (\vvec{\ell}_{j'} - \vvec{\ell}_j) / \|\vvec{\ell}_{j'} - \vvec{\ell}_j\|$. 
In words, $f_\mathrm{Smooth}$ encourages landmarks to be locally smooth with respect to the reference landmark's surface normal ($\vvec{n}_j$) by enforcing that $\vvec{d}_{j',j}$, i.e., the vector pointing from the reference landmark ($\vvec{\ell}_j$) towards a neighboring landmark ($\vvec{\ell}_{j'}$), be perpendicular to $\vvec{n}_j$. 
We will demonstrate that adding these smoothness constraints will result in more feasible surface normal estimates and lower photometric errors.


\section{Experimental Setup} \label{sec:experiments}
\vspace{5pt}
\begin{figure}[tb!]
    \centering
    \begin{subfigure}[t]{0.46\linewidth}
      \includegraphics[width=\linewidth]{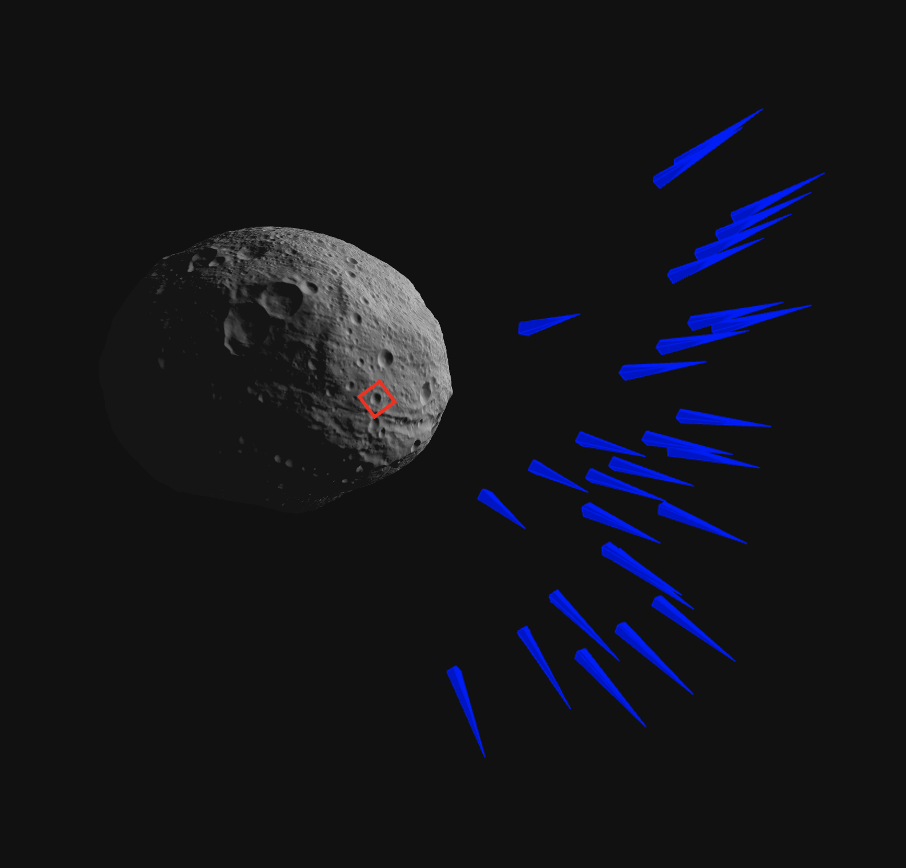}
    \end{subfigure}%
    \hfill
    \begin{subfigure}[t]{0.53\linewidth}
      \includegraphics[width=\linewidth]{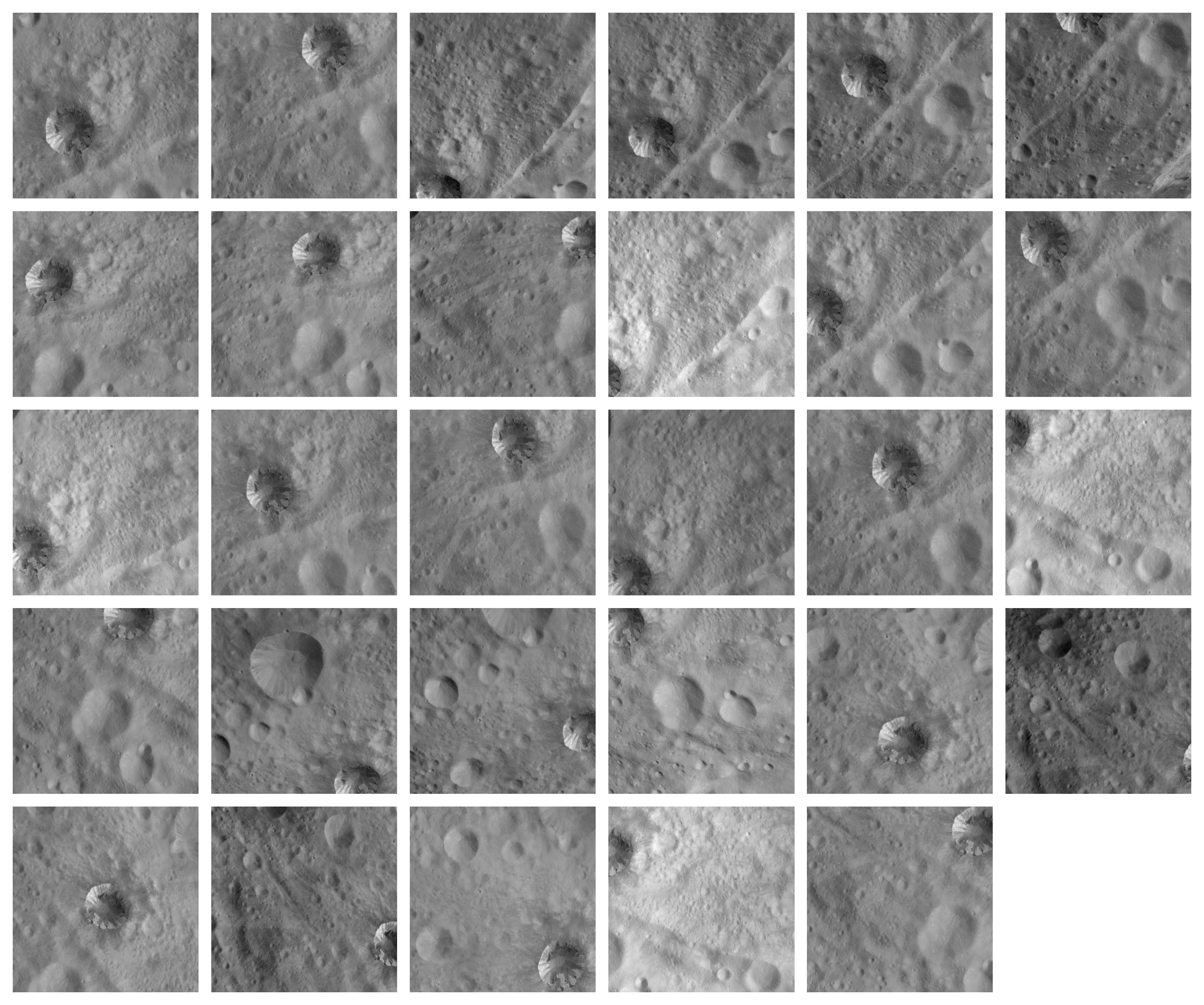}
    \end{subfigure}
    \caption{Image sequence used in this investigation. Camera poses and \textit{global} shape model are drawn according to the SPC solution.}
    \label{fig:sequence-noaxis}
\end{figure}

Our experiments leverage imagery of the Cornelia crater on Asteroid 4 Vesta captured by NASA's Dawn mission~\cite{russell2012dawn,sierks2011dawn} for evaluation of the proposed approach. 
These images are publicly available on NASA's Planetary Data System (PDS)~\cite{pds} and maintained by NASA's Navigation and Ancillary Information Facility (NAIF), each with a resolution of $1024 \times 1024$ pixels. 
Cornelia is an approximately 15 km in diameter crater located at 15.57$^\circ$E and 9.37$^\circ$S (in the Claudia double-prime coordinate frame~\cite{li2012claudia}) and has been the subject of numerous studies~\cite{denevi2012pitted,williams2014lobate,schroder2013resolved}.  
Cornelia was also chosen in part due to its interesting albedo features, which represent a challenging scenario for photometric reconstruction methods such as SPC and the proposed approach.
The full image sequence used in this investigation, captured during the High Altitude Mapping Orbit (HAMO) phase, is visualized in Figure \ref{fig:sequence-noaxis}, which features 29 images captured at a mean orbital radius of $\sim$$950$ km and a ground sample distance of approximately 60 m per pixel.


\subsection{Experimental Baselines} \label{sec:baselines}

\begin{figure}
    \centering
    \input{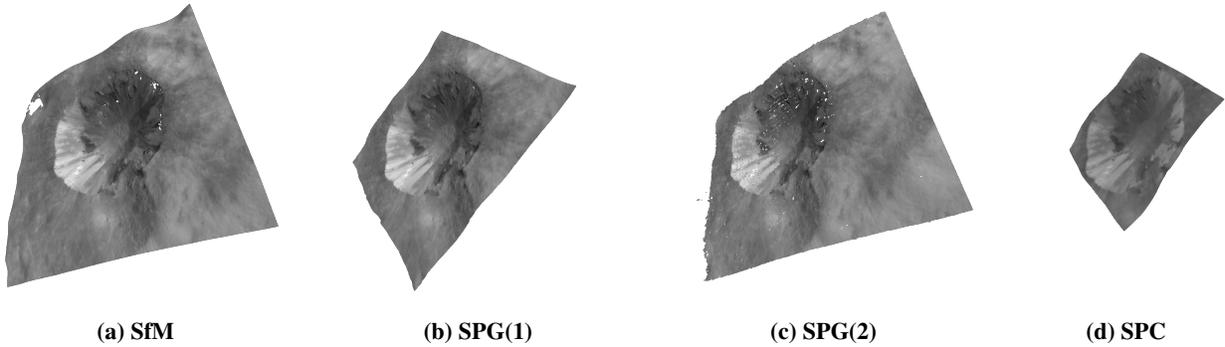}
    \caption{Baselines for comparison.}
    \label{fig:baselines}
\end{figure}

We compare our pipeline against four different baselines (see Figure \ref{fig:baselines}), including the SPC solution and two different methods based on stereophotogrammetry (SPG):

\paragraph{SPC} 
An important point of comparison is with respect to the traditional SPC~\cite{gaskell2008,gaskell2023psj} pipeline, which will provide a baseline for our surface normal and albedo estimates, along with pose and landmark position values. 
The SPC reconstruction leverages the Schr\"{o}der reflectance model as opposed to the nominal Lunar-Lambert function. 
A bigmap of the Cornelia region was created by merging 289 overlapping sub-maps, or maplets. 
The maplets, and the resulting bigmap, have a spatial resolution of 30 meters. 
Approximately 50 pictures were used to model the terrain, captured during both the High Altitude Mapping Orbit (HAMO) and Low Altitude Mapping Orbit (LAMO) phases. 
The HAMO images, which constituted most of the imagery, had a ground sample distance (GSD) of approximately 60 meters per pixel, while the images from the LAMO phase featured a GSD of 20 meters per pixel. 
Not all pictures were used in every maplet.  
\textit{A priori} pose information was sampled from a global orbital dynamics solution developed for the purpose of gravity science.
\textit{A priori} height information for the maplets was sampled from a previously converged set of maps at a spatial resolution of 50 meters, which were in turn derived from 100 meter spatial resolution maps.
More details about the SPC reconstruction process at Vesta can be found in \cite{mastrodemos2012}.

\paragraph{SPG(1)} 
The first method based on SPG leverages multi-baseline stereo matching and triangulation. 
The pose estimates from the SPC solution were used, and a reference image was chosen (the first image in Figure \ref{fig:sequence-noaxis}) to find the pixel correspondents from the rest of images. 
Because the image poses from SPC are very accurate, it allowed for precise epipolar rectification and 1D disparity search. 
To improve the matching robustness, the images were also normalized~\cite{briechle2001template}. 
Then, the 3D position of each point is triangulated~\cite[Chapter~4.1]{hartley2003multiple} to obtain the final map.

\paragraph{SPG(2)} 
A second multi-baseline stereo approach was also used. 
Again, the pose estimates from SPC were leveraged. 
Rather than using a reference image, all images were rectified to a series of densely spaced, parallel planes that sweep through the volume of interest. 
Then, given a point in any image, its 3D location is determined by the rectifying plane which maximized photo-consistency across the entire dataset~\cite{xiong2005}.

\paragraph{SfM} This is the dense SfM solution, i.e., the the SPC-SfM process without the $f_\mathrm{SPC}$, $f_\mathrm{SS}$, and $f_\mathrm{Smooth}$ factors. 

\paragraph{} Since our method does not assume any priors on the camera poses or landmark positions, resulting in a scale ambiguity, we must align our solution to the baselines before comparison. 
To do this, we estimate a $\mathrm{Sim}(3)$ transformation between the estimated and ground truth camera poses and a subset of 180 of the landmarks, by Karcher mean and \cite{Zinsser2005icprip} (implemented in GTSAM's \texttt{Similarity3.align} function), and apply this transformation to our reconstructed poses, landmarks, and surface normals. 
Since the SPG solutions use the poses estimated by SPC, we align to the SPC solution before comparing to the SPG(1) and SPG(2) solutions. 


\subsection{Implementation Details}

Keypoint measurements and matches are computed using DKM, which provides dense, per-pixel correspondences, where the first image (the top left image in Figure \ref{fig:sequence-noaxis}) is taken to be the reference image, as described in Section \ref{sec:gtsfm}. 
The matching is constrained to a $500 \times 500$ pixel region, drawn in red in Figure \ref{fig:sequence-noaxis}, as this contains the extent of the baseline SPC and SPG maps. 
Thus, each of the maps estimated by our approach contains approximately 250,000 points. 
The keypoint measurements are assigned a covariance of $\Sigma_{j,k} = \mathrm{I}_2$.
Image brightness values are (bilinearly) interpolated at the keypoint measurements $\Hat{\vvec{p}}_{j,k}$ to derive the measurements $\Hat{I}_k(\Hat{\vvec{p}}_{j,k})$ used in the proposed SPC factors $f_\mathrm{SPC}$ defined in Equation \eqref{eq:fSPC}. 
The image brightness measurements are assigned a standard deviation of $\sigma_I = 0.5$ for the uncalibrated case and $\sigma_I = 0.01$ for the calibrated case, which we found to work well empirically. 
Only landmarks with $\geq 8$ keypoint measurements are inserted into the graph. 

The (simulated) Sun sensor measurements $\hat{\vvec{s}}^\fC_k \in \mathbb{S}^2$ were derived from the normalized ground truth values from SPC of the Sun's relative position to the origin of the camera frame $\oC_k$ expressed in the camera frame $\fC_k$, i.e., $\vvec{r}_{\oI\oC_k}^{\fC_k}$, and assigned an uncertainty of $\Sigma_\xi =  \sigma_\xi^2\mathrm{I}_2$ where $\sigma_\xi = 1 \times 10^{-3}$. 
Next, the surface normals are initialized by finding the $32$ closest neighbors to each point in the point cloud and fitting a plane to this local terrain, and the normal to the plane is taken as the initial surface normal. 
These initial surface normals are then used to initialize the albedo by independently computing the albedo in each image using the initial camera poses and landmark positions, where the initial albedo of each landmark is taken to be the average albedo computed over all views from which it was seen.

The smoothness factors $f_\mathrm{Smooth}$ (Equation \eqref{eq:fsmooth}) are inserted into the graph according to the keypoints in the reference image (corresponding to $k=0$). 
Specifically, a smoothness factor is inserted between a landmark, $\vvec{\ell}_j$ (and its associated surface normal, $\vvec{n}_j$) and a neighboring landmark, $\vvec{\ell}_{j'}$, if $\|\Hat{\vvec{p}}_{j,0} - \Hat{\vvec{p}}_{j',0}\| \leq 1$. 
Thus, a smoothness factor will be inserted between a landmark and at most four other landmarks in the map since DKM computes \textit{per-pixel} matches between an image pair.
We found a very small value for the local smoothness weight to work well for our experiment, where we used a value of $\eta = 10^{-4}$.
As previously discussed, we leverage the GTSAM library~\cite{dellaert2012} to model the proposed keypoint-based SPC problem using factor graphs and optimize the resulting nonlinear least-squares using the Levenberg-Marquardt algorithm and the analytical partial derivatives of the measurement functions for the respective factors (see Appendix \hyperref[sec:partials]{A}). 


\subsection{Performance metrics} \label{sec:metrics}

We define the following metrics to measure the performance of the proposed approach:
\begin{equation}
    \delta \ell_j \triangleq \| \bm{\ell}_j - \overline{\bm{\ell}}_j\|_2,
\end{equation}
\begin{equation}
    \delta\epsilon_j \triangleq \cos^{-1}\left(\mathbf{n}_j^\top\overline{\mathbf{n}}_j\right),
\end{equation}
and
\begin{equation}
    \delta a_j \triangleq |a_j - \overline{a}_j| / \overline{a}_j.
\end{equation}
As before, $\overline{\bm{\ell}}$, $\overline{\bm{n}}$, $\overline{a}$ denote the ground truth values of the landmark position, surface normal, and albedo, respectively. 
These ground truth values are assigned by finding the closest point in our reconstructed map to that of the baseline (after the $\mathrm{Sim}(3)$ alignment step), and taking the position of that landmark, as well as the associated normal and albedo for the SPC baseline, as the ground truth. 
For quantitative evaluation of the estimated trajectory, for each time index $k= 0,\ldots,n$, we compute the error 
\begin{align}
    \begin{bmatrix}\delta \vvec{\kappa}_k^\top & \delta \vvec{r}_k^\top\end{bmatrix} \triangleq  \desk{\log \left(T_{\fB\fC_k}^{-1}\overline{T}_{\fB\fC_k}\right)} \in \mathbb{R}^6,
\end{align}
between the ground truth relative pose $\overline{T}_{\fB\fC_k}$ and the estimated pose $T_{\fB\fC_k}$, where $\log : \mathrm{SE}(3) \rightarrow \mathfrak{se}(3)$ is the $\mathrm{SE}(3)$ logarithm map at the identity and $\vee$ is the \textit{vee} operator, as detailed in \cite{chirikjian2011}, which gives
\begin{align}
    \|\delta\vvec{\kappa}_k\| &= \cos^{-1}\left(\frac{\mathrm{trace}(R_{\fC_k\fB}\overline{R}_{\fB\fC_k}) - 1}{2}\right), \\
    \|\delta\vvec{r}_k\| &= \|R_{\fC_k\fB}\overline{\vvec{r}}_{\oC_k\oB}^{\fB} + \vvec{r}_{\oB\oC_k}^{\fC_k}\|.
\end{align}
Finally, the root mean squared error between the measured, $\Hat{I}_k(\Hat{\vvec{p}}_{j,k})$, and estimated, $I(T_k, \mathbf{s}_k, \vvec{\ell}_j, \vvec{n}_j, a_j)$, image brightness values, normalized by the average measured brightness, is taken to be the photometric error, as in \cite{schroder2013resolved}:
\begin{equation}
    \delta I_j \triangleq \left(\frac{1}{|\mathcal{K}_j|}\sum_{k\in \mathcal{K}_j} \Hat{I}_k(\Hat{\vvec{p}}_{j,k})\right)^{-1}\sqrt{\frac{1}{|\mathcal{K}_j|}\sum_{k\in \mathcal{K}_j}\left(I(T_k, \mathbf{s}_k, \vvec{\ell}_j, \vvec{n}_j, a_j) - \Hat{I}_k(\Hat{\vvec{p}}_{j,k})\right)^2},
\end{equation}
where $\mathcal{K}_j$ denotes the set of indices of the images from which the $j^{th}$ landmark was viewed.


\section{Results} \label{sec:results}
\vspace{5pt}
\begin{table*}[tb!]
\footnotesize
\centering
\ra{1.5}
\caption{Mean values of the performance metrics (outlined in Section \ref{sec:metrics}) with respect to each baseline. Pose errors were not computed for the SPG solutions, as they leverage the pose estimates from SPC (see Section \ref{sec:baselines}).}
\begin{adjustbox}{width=\linewidth}
\begin{tabular}{lrrrcrrrcrrrcrrrrrr}\toprule
 & \multicolumn{3}{c}{SfM} && \multicolumn{3}{c}{SPG(1)} && \multicolumn{3}{c}{SPG(2)} && \multicolumn{6}{c}{SPC} \\
\cmidrule{2-4} \cmidrule{6-8} \cmidrule{10-12} \cmidrule{14-19}
Model & $\delta\ell\ \left[\text{km}\right]$ & $\|\delta\vvec{r}\|\ \left[\text{km}\right]$ & $\|\delta\bm{\kappa}\|\ \left[^{\circ}\right]$ && $\delta\ell\ \left[\text{km}\right]$ & $\|\delta\vvec{r}\|\ \left[\text{km}\right]$ & $\|\delta\bm{\kappa}\|\ \left[^{\circ}\right]$ && $\delta\ell\ \left[\text{km}\right]$ & $\|\delta\vvec{r}\|\ \left[\text{km}\right]$ & $\|\delta\bm{\kappa}\|\ \left[^{\circ}\right]$ && $\delta\ell\ \left[\text{km}\right]$ & $\|\delta\vvec{r}\|\ \left[\text{km}\right]$ & $\|\delta\vvec{\kappa}\|\ \left[^{\circ}\right]$ & $\delta\epsilon_\mathrm{s}\ \left[^{\circ}\right]$ & $\delta\epsilon_\mathrm{n}\ \left[^{\circ}\right]$ & $\delta a\ \left[\%\right]$ \\ 
\midrule
\rowcolor[gray]{0.9}
Lunar-Lambert                    & & & && & & && & & && & & & & & \\
\rowcolor[gray]{0.9}
$\quad$ \textit{Uncalibrated}                 &  0.0017 & 0.0882 & 0.0082 && 0.0425 & --- & --- && 0.0563 & --- & --- && 0.0881 & 1.0850 & 0.0966 & 0.6061 & 13.72 & 6.05\\
\rowcolor[gray]{0.9}
$\quad$ \textit{Uncal}. $+ f_\mathrm{smooth}$ &  0.0048 & 0.0904 & 0.0080 && 0.0419 & --- & --- && 0.0579 & --- & --- && 0.0867 & 1.0879 & 0.0966 & 0.5743 & 4.80 & 3.85\\
\rowcolor[gray]{0.9}
$\quad$ \textit{Cal}. $+ f_\mathrm{smooth}$   &  0.0209 & 1.3522 & 0.1497 && 0.0507 & --- & --- && 0.0626 & --- & --- && 0.0822 & 1.9542 & 0.1988 & 4.8529 & 7.92 & 6.96\\
\midrule
Sch\"{o}der                  & & & && & & && & & & & & \\
$\quad$ \textit{Uncalibrated}                 &  0.0037 & 0.1589 & 0.0159 && 0.0408 & --- & --- && 0.0576 & --- & --- && 0.0885 & 1.0985 & 0.0987 & 0.6222 & 13.40 & 4.31\\
$\quad$ \textit{Uncal}. $+ f_\mathrm{smooth}$ &  0.0052 & 0.1285 & 0.0135 && 0.0432 & --- & --- && 0.0582 & --- & --- && 0.0867 & 1.0850 & 0.0977 & 0.5389 & 4.80 & 3.85\\
$\quad$ \textit{Cal}. $+ f_\mathrm{smooth}$   &  0.0057 & 0.1977 & 0.0226 && 0.0433 & --- & --- && 0.0585 & --- & --- && 0.0867 & 1.0965 & 0.1013 & 0.5170 & 4.78 & 3.85\\
\bottomrule
\end{tabular}
\end{adjustbox}\\
\label{tab:spc-results}
\end{table*}

\begin{figure}[ht!]
    \centering
    \input{figs/meas-est-zoom}
    \caption{Per image photometric error and qualitative comparison between measured and estimated brightness values for the Schr\"{o}der (\textit{uncalibrated}) model with smoothness constraints. The legend in (a) is formatted according to $k: (\lambda_k, \xi_k)$ where the biases $\xi_k$ have been normalized by the average brightness of each image.}
    \label{fig:meas-est-zoom}
\end{figure}

The resulting albedo map and surface normals, along with the photometric errors, from our proposed SPC-SfM solution are provided in Figures \ref{fig:reconstruction-results-llambert} and \ref{fig:reconstruction-results-schroder} for the Lunar-Lambert and Schr\"{o}der models, respectively.
For the \textit{uncalibrated} cases, we see that our method is able to reconstruct maps with very low photometric errors of 0.0147 and 0.0122 for the Lunar-Lambert and Schr\"{o}der model, respectively, albeit with numerous artifacts in the surface normal maps. 
The addition of the smoothness constraints is shown to generate coherent surface normals while also significantly reducing the average photometric error and bringing the photometric error for the Lunar-Lambert model in line with the Schr\"{o}der model at 0.0108. 
These extremely low photometric errors result in renderings that are virtually indistinguishable from the captured images, as shown in Figure \ref{fig:meas-est-zoom}. 
However, the \textit{calibrated} case paints a slightly different picture in terms of the fidelity of the examined reflectance functions, where the Lunar-Lambert model results in an average photometric error of 0.0227, over twice that of the Schr\"{o}der model on the same data. 
This may be unsurprising, as the Lunar-Lambert equation was fit to images of the Moon while the Schr\"{o}der model leveraged approach imagery of Vesta.
However, this phenomenon has not been demonstrated in previous works on SPC that leverage only uncalibrated imagery, where these works indicate that the choice of reflectance model has little effect on the quality of the final DTM~\cite{gaskell2023psj}.  
Moreover, this suggests that the scale and bias variables estimated in the uncalibrated case are able to ``explain away'' many of the sources of error due to the underlying reflectance model and motivates the estimation and use of specialized reflectance models for different small bodies, especially when dealing with radiometrically calibrated images. 

Next, we compare our solution to the SPC reconstruction in Figures \ref{fig:spc-comparison-llambert} and \ref{fig:spc-comparison-schroeder} for the Lunar-Lambert and Schr\"{o}der models, respectively.
Again, we see that the Lunar-Lambert model with the radiometrically calibrated imagery results in higher errors as compared to the other solutions. 
The uncalibrated cases without the smoothness factors display surface normal errors of $\sim$$13^\circ$ with respect to SPC, but these errors are virtually rectified with the addition of the smoothness factors, which bring the average error to $<$$5^\circ$. 
The relative albedo errors are also exceedingly low, almost all of which are below 4\%, except for the calibrated case for the Lunar-Lambert model and the uncalibrated cases without smoothness constraints, which are still below 7\% on average. 
However, all solutions exhibit regions with relatively large errors with respect to the landmark positions (on the order of 350 m). 
Yet, these large landmark error regions are not present when comparing against the SfM and SPG solutions, as shown in Figures \ref{fig:spc-sfm-spg-comparison-llambert} and \ref{fig:spc-sfm-spg-comparison-schroder}. 
This suggests that the errors in the landmarks arise from errors in the SPC solution, as opposed to errors in our approach, purportedly due to errors in the associated surface normal, which seem to originate at the dark albedo feature near the northwest edge of the crater. 
Indeed, since the landmark heights in the SPC solution are computed by integrating the slopes~\cite{mastrodemos2012}, errors in the slope translate to errors in the landmark positions that propagate from the points in the map where the slope errors arise towards the direction of the integration. 
Conversely, since our landmarks are not explicitly derived from the surface normal estimates, and instead are independently estimated and constrained by the keypoint measurements, our pipeline does not seem to be as susceptible to surface normal errors as the traditional SPC pipeline. 

These results, along with the associated pose errors with respect to the SPC and SfM solutions, are also summarized in Table \ref{tab:spc-results}. 
Except for the Lunar-Lambert model on calibrated imagery, all position errors and orientation errors are approximately 1 km and $0.1^\circ$, respectively, as compared to the SPC solution. 
Thus, we achieve positions errors of $\sim$$0.1\%$ relative to the mean orbital radius of 950 km without requiring \textit{a priori} estimates of the camera's pose or an initial topography solution. 
Moreover, our pipeline also refines the Sun-relative direction, where our Sun vector estimates are within $0.6^\circ$ of the SPC estimates. 
When comparing against the SfM solution, our position and orientation estimates align even better, with position errors of $<0.2$ km ($<0.1$ km) and orientation errors of $<0.025^\circ$ ($<0.01^\circ$) for the Schr\"{o}der (Lunar-Lambert) model. 
It is unclear why the Lunar-Lambert SPC-SfM solution seems to demonstrate marginally lower pose errors than the Schr\"{o}der model as compared to the SfM solution. 
In any case, the position errors for the Schr\"{o}der model as compared to the SfM solution are $\sim$$0.02\%$ relative to the mean orbital radius. 
The Lunar-Lambert model on calibrated imagery exhibits inflated errors with respect to almost all metrics, demonstrating how incorrectly modeled reflectance properties can perturb the solution. 

\afterpage{%
    \clearpage
    \begin{landscape}
    \begin{figure}[p]
    \centering
    \begin{minipage}{.48\linewidth}
    \input{figs/reconstruction-results-llambert}
    \caption{Reconstructed surface normal and albedo maps, along with the photometric errors, for the Lunar-Lambert reflectance function. The (relative) albedos for the uncalibrated case are normalized by the maximum albedo in the map.}
    \label{fig:reconstruction-results-llambert}
    \end{minipage}%
    \hfill
    \begin{minipage}{.48\linewidth}
    \input{figs/reconstruction-results-schroder}
    \caption{Reconstructed surface normal and albedo maps, along with the photometric errors, for the Schr\"{o}der reflectance function. The (relative) albedos for the uncalibrated case are normalized by the maximum albedo in the map.}
    \label{fig:reconstruction-results-schroder}
    \end{minipage}
    \end{figure}
    \end{landscape}
}

\afterpage{%
    \clearpage
    \begin{landscape}
    \begin{figure}[p]
    \centering
    \begin{minipage}{.48\linewidth}
    \input{figs/spc-comparison-llambert}
    \caption{Quantitative comparison between our solution and the SPC solution with respect to landmark error, normal error, and relative albedo error for the Lunar-Lambert reflectance function.}
    \label{fig:spc-comparison-llambert}
    \end{minipage}%
    \hfill
    \begin{minipage}{.48\linewidth}
    \input{figs/spc-comparison-schroeder}
    \caption{Quantitative comparison between our solution and the SPC solution with respect to landmark error, normal error, and relative albedo error for the Schr\"{o}der reflectance function.}
    \label{fig:spc-comparison-schroeder}
    \end{minipage}
    \end{figure}
    \end{landscape}
}

\afterpage{%
    \clearpage
    \begin{landscape}
    \begin{figure}[p]
    \centering
    \begin{minipage}{.48\linewidth}
    \input{figs/spc-sfm-spg-comparison-llambert}
    \caption{Quantitative comparison between our solution and the SPC, SfM, and SPG solutions with respect to landmark error for the Lunar-Lambert model.}
    \label{fig:spc-sfm-spg-comparison-llambert}
    \end{minipage}%
    \hfill
    \begin{minipage}{.48\linewidth}
    \input{figs/spc-sfm-spg-comparison-schroder}
    \caption{Quantitative comparison between our solution and the SPC, SfM, and SPG solutions with respect to landmark error for the Schr\"{o}der model.}
    \label{fig:spc-sfm-spg-comparison-schroder}
    \end{minipage}
    \end{figure}
    \end{landscape}
}


\section{Conclusion}

This paper proposes the incorporation of photometric stereo constraints leveraged by stereophotoclinometry (SPC) into a keypoint-based structure-from-motion (SfM) system to estimate the surface normal and albedo at landmarks to improve surface and shape characterization of small celestial bodies from \textit{in-situ} imagery. 
The proposed framework was validated on real imagery of the Cornelia crater on Asteroid 4 Vesta using dense keypoint measurements and correspondences from a deep learning-based method, i.e., DKM~\cite{edstedt2023dkm}. 
The results demonstrate that the proposed framework is able to estimate accurate surface normal and albedo estimates as compared to the baseline SPC values.
Moreover, our approach does not rely on high-fidelity \textit{a priori} estimates of the camera's pose or an inital topography solution, and can operate \textit{autonomously} thanks to the accurate keypoint correspondences from the front-end. 
While the proposed approach assumes a global reflectance function (and phase function for the Schr\"{o}der model), this assumption may not be sufficient for more heterogenous bodies such as Asteroid 101955 Bennu~\cite{golish2021}.
Future work will include continued validation and analysis of the proposed approach on other celestial bodies (e.g., the Moon, Bennu) and incorporating reflectance function parameter estimation. 
Our reconstructed maps and camera poses will be made available at \href{https://github.com/travisdriver/spc-factor-results}{\texttt{https://github.com/travisdriver/spc-factor-results}}. 


\section*{Appendix}
\vspace{5pt}
\section*{A. Measurement Function Partial Derivatives} \label{sec:partials}

In this section, we explicitly define the partial derivatives for the measurement function $I(T, \mathbf{s}, \vvec{\ell}, \vvec{n}, a)$, with the Lunar-Lambert model (Equation \eqref{eq:ref-ll}), for the factor $f_\mathrm{SPC}$ (Equation \eqref{eq:fSPC}), which are used by GTSAM~\cite{dellaert2012} to solve the nonlinear least-squares problem defined in Equation \eqref{eq:logf}.
The nonlinear least-squares objective is reparameterized according to the outlined retractions, i.e., Equations \eqref{eq:ret-pose} and \eqref{eq:ret-unit}. 
This reparameterization is referred to as \textit{lifting}~\cite{absil2007trust} and frames the optimization problem in terms of coordinates in the tangent space of the variables.
Recall that $\vvec{e} = \vvec{r}_{\oC\oB} - \vvec{\ell}$. 
Then,
\begin{equation}
    \frac{\partial \vvec{e}}{\partial \vvec{\zeta}} = \begin{bmatrix}
        \mathrm{I}_{3\times 3} & R
    \end{bmatrix}, \qquad
    \frac{\partial \vvec{e}}{\partial \vvec{\ell}} = -\mathrm{I}_{3\times 3},
\end{equation}
where we have applied the retraction $r_{\oC\oB} \leftarrow r_{\oC\oB} + R\vvec{\tau}$. 
Let $\vvec{d} = \vvec{e}/\|\vvec{e}\|$, $f = \vvec{s}^\top\vvec{n}$, $w = \vvec{d}^\top\vvec{n}$, and $h = \vvec{s}^\top\vvec{d}$. Then 
\begin{equation}
    \begin{split}
        \frac{\partial f}{\partial \vvec{\xi}_{\vvec{s}}} &= \vvec{n}^\top B_{\vvec{s}},  \\
        \frac{\partial w}{\partial \vvec{\xi}_{\vvec{d}}} &= \vvec{n}^\top B_{\vvec{d}},  \\
        \frac{\partial h}{\partial \vvec{\xi}_{\vvec{s}}} &= \vvec{d}^\top B_{\vvec{s}},  \\
        \frac{\partial \vvec{\xi}_{\vvec{d}}}{\partial \vvec{e}}& = B_{\vvec{d}}^\top D_\mathrm{norm}(\vvec{e}),
    \end{split}
    \qquad
    \begin{split}
        \frac{\partial f}{\partial \vvec{\xi}_{\vvec{n}}} &= \vvec{s}^\top B_{\vvec{n}}, \\
        \frac{\partial w}{\partial \vvec{\xi}_{\vvec{n}}} &= \vvec{d}^\top B_{\vvec{n}}, \\
        \frac{\partial h}{\partial \vvec{\xi}_{\vvec{d}}} &= \vvec{s}^\top B_{\vvec{d}},
    \end{split}
\end{equation}
where we have applied the retraction $\vvec{x} \leftarrow \vvec{x} + B_{\vvec{x}}\vvec{\xi}_{\vvec{x}}$ with $B_{\vvec{x}}$ and $\vvec{\xi}_{\vvec{x}}$ representing the basis for and local coordinates in the tangent space at a unit vector $\vvec{x} \in \mathbb{S}^2$, respectively, and
\begin{equation}
    D_\mathrm{norm}(\vvec{v}) =
    \begin{bmatrix}
        v_y^2 + v_z^2 & -v_xv_y & -v_xv_z \\
        -v_xv_y & v_x^2 + v_z^2 & -v_yv_z \\
        -v_xv_z & -v_yv_z & v_x^2 + v_y^2
    \end{bmatrix} (v_x^2 + v_y^2 + v_z^2)^{-3/2}.
\end{equation}
Next, let $b = -\cos^{-1}(h) / 60$ and $g = \exp(b)$. Then,
\begin{equation}
    \begin{split}
        \frac{\partial b}{\partial h} &= \left(60\sqrt{1 - h^2}\right)^{-1}, \\
        \frac{\partial g}{\partial b} &= \exp(b), \\
    \end{split}
    \qquad
    \begin{split}
        \frac{\partial g}{\partial \vvec{\xi}_{\vvec{s}}} &= \frac{\partial g}{\partial b}\frac{\partial b}{\partial h}\frac{\partial h}{\partial \vvec{\xi}_{\vvec{s}}}, \\
        \frac{\partial g}{\partial \vvec{\xi}_{\vvec{d}}} &= \frac{\partial g}{\partial b}\frac{\partial b}{\partial h}\frac{\partial h}{\partial \vvec{\xi}_{\vvec{d}}}.
    \end{split}
\end{equation}
This gives
\begin{equation}
    \frac{\partial I}{\partial \vvec{e}} = a \left(   -f\frac{\partial g}{\partial \vvec{\xi}_{\vvec{d}}} + 2f \left( \frac{1}{f + w} \frac{\partial g}{\partial \vvec{\xi}_{\vvec{d}}} - \frac{g}{(f + w)^2} \frac{\partial w}{\partial \vvec{\xi}_{\vvec{d}}} \right)   \right)\frac{\partial \vvec{\xi}_{\vvec{d}}}{\partial \vvec{e}}.
\end{equation}
Finally,
\begin{align}
    \frac{\partial I}{\partial \zeta} &= \frac{\partial I}{\partial\vvec{e}} \frac{\partial\vvec{e}}{\partial \zeta}, \\
    \frac{\partial I}{\partial \vvec{\xi}_{\vvec{s}}} &= a \left(   \frac{\partial f}{\partial \vvec{\xi}_{\vvec{s}}} - \left( \frac{\partial g}{\partial \vvec{\xi}_{\vvec{s}}}f + g \frac{\partial f}{\partial \vvec{\xi}_{\vvec{s}}} \right) + \frac{2}{f + w}\left(\frac{\partial g}{\partial \vvec{\xi}_{\vvec{s}}}f + g \frac{\partial f}{\partial \vvec{\xi}_{\vvec{s}}}\right) - \frac{2gf}{(f + w)^2}\frac{\partial f}{\partial \vvec{\xi}_{\vvec{s}}}  \right), \\
    \frac{\partial I}{\partial \vvec{\ell}} &= \frac{\partial I}{\partial\vvec{e}} \frac{\partial \vvec{e}}{\partial\vvec{\ell}}, \\
    \frac{\partial I}{\partial \vvec{\xi}_{\vvec{n}}} &= a \left((1 - g)\frac{\partial f}{\partial \vvec{\xi}_{\vvec{n}}} + 2g \left(\frac{1}{f + w}\frac{\partial f}{\partial \vvec{\xi}_{\vvec{n}}} - \frac{f}{(f + w)^2}\left(\frac{\partial f}{\partial \vvec{\xi}_{\vvec{n}}} + \frac{\partial w}{\partial \vvec{\xi}_{\vvec{n}}}\right)\right)   \right), \\
    \frac{\partial I}{\partial a} &= (1 - g) f + g \frac{2f}{f + w}.
\end{align}


\section*{Acknowledgments}

This work was supported by NASA Space Technology Graduate Research Opportunity   80NSSC21K1265.
A portion of this research was carried out at the Jet Propulsion Laboratory, California Institute of Technology, under a contract with the National Aeronautics and Space Administration (80NM0018D0004). 
The authors would like to thank Kenneth Getzandanner and Andrew Liounis from NASA Goddard Space Flight Center for several helpful discussions and comments.
We also thank Stefanus E. Schr\"{o}der from Lule\r{a} University of Technology for providing support for the radiometric calibration process~\cite{schroder2013cal,schroder2014cal}.

\bibliography{sample}

\end{document}